\documentclass[a4paper,11pt,oneside]{mythesis}
\usepackage{fix-cm}
\usepackage{amsmath}
\usepackage{amssymb}
\usepackage[normalem]{ulem}
\usepackage{algorithm}
\usepackage{algpseudocode}
\usepackage{listings}
\usepackage{hyperref}
\usepackage{caption}
\usepackage{subcaption}
\usepackage{multirow}
\usepackage{lmodern}

\title{Weight Averaging for \\Out-of-Distribution Generalization and \\ Few-Shot Domain Adaptation}
\typethesis{MSc thesis}  
\author{Shijian Xu}
\supervisor{Dr. Grigorios Chrysos}
\adviser{Prof. Volkan Cevher}
\expert{Dr. Stratis Tzoumas}

\newcommand{\dtr}{{d_\mathrm{tr}}}

\begin{document}
\frontmatter
\pagenumbering{roman}
\titlepage

\begin{abstract}

Empirical risk minimization (ERM) is not robust to changes in the distribution of data. When the distribution of test data is different from that of training data, the problem is known as out-of-distribution generalization. Recently, two techniques have been developed for addressing out-of-distribution generalization in computer vision: weight averaging (WA) and sharpness-aware minimization (SAM). WA involves training multiple models with different hyperparameters and then averaging the weights of these models, which can significantly improve out-of-distribution generalization performance. SAM optimizes a neural network to find minima in flat regions, which have been proven to perform well under distribution shifts. While these techniques have made great progress, there is still room for improvement and further exploration. In this thesis, we propose increasing the model diversity in WA explicitly by introducing gradient similarity as a loss regularizer to further improve out-of-distribution generalization performance. We also propose combining WA and SAM to solve the problem of few-shot domain adaptation. Our extensive experiments on digits datasets (MNIST, SVHN, USPS, MNIST-M) and other domain adaptation datasets (VLCS, PACS) show that combining WA and SAM leads to improved out-of-distribution generalization performance and significantly increases few-shot domain adaptation accuracy.

\end{abstract}
\clearpage

\section*{Acknowledgements}

It was a great experience to conduct my master thesis at LIONS lab.
I would like to express my deepest gratitude to my thesis supervisor, Dr. Grigorios Chrysos, for his invaluable guidance, support, and encouragement throughout the entire research and writing process. 
His expertise and dedication to my work have been instrumental in the completion of this thesis.
I also extend my thanks to the members of my thesis committee, Prof. Volkan Cevher and Dr. Stratis Tzoumas, for their time, effort, and valuable feedback in reviewing and evaluating my work.
I am grateful to my colleagues and friends who provided support and inspiration throughout my research journey, especially Zhengyu Zhu, and Yongtao Wu.
Lastly, I would like to acknowledge my family for their unwavering love, support, and understanding, especially during the challenging times throughout this process.
This thesis would not have been possible without the help and support of all of these individuals. \clearpage
\tableofcontents* \clearpage

\mainmatter
\pagestyle{TUDelft}\pagenumbering{arabic}
\counterwithout{table}{chapter}
\counterwithout{figure}{chapter}

\section{Introduction}
\label{sec:thesis_introduction}  



In recent years, significant advancements in machine learning have been made possible through the use of large pre-trained models like BERT~\cite{devlin2018bert}, GPT-3~\cite{brown2020language}, CLIP~\cite{radford2021learning}, and DDPM~\cite{ho2020denoising}.

It has also been found that transferring models pre-trained on large-scale datasets can lead to improved performance and sample efficiency~\cite{kolesnikov2020big}. As a result, pre-training on large datasets and fine-tuning the model on downstream tasks or target datasets has become a common practice for various computer vision tasks, including domain generalization and domain adaptation.

However, these advancements in machine learning have been hindered by a phenomenon known as distribution shift~\cite{nagarajan2020understanding}. The success of pre-trained models, such as those mentioned above, relies on the assumption that the source (train) data and the target (test) data have the same distribution, known as being independent and identically distributed (i.i.d.). When this assumption is not met, the performance of these models may suffer. This is particularly concerning in fields where the consequences of poor performance could be severe, such as digital health and medical AI. In these areas, it is essential that machine learning models maintain a high level of accuracy, as even small performance degradation could have serious consequences. Therefore, addressing and mitigating the effects of distribution shift is of the utmost importance in the development and deployment of machine learning models.

In this paper, we focus on two critical research topics related to distribution shift: out-of-distribution (OOD) generalization and few-shot domain adaptation. OOD generalization refers to the ability of a machine learning model to accurately predict on data that comes from a distribution different from the one it was trained on. This is an important consideration, as real-world data often experiences distribution shifts, where the distribution of the training data differs from the distribution of the test data. The ability to generalize to new, OOD data is essential for the long-term effectiveness and deployment of machine learning models, particularly as they are applied to a wider range of tasks and environments. 
However, OOD generalization remains a challenging problem.
The challenges come from multiple perspectives. 
First, OOD generalization is too general and there are various downstream tasks can be considered into OOD generalization.
Second, the extrapolation ability of neural networks is questioned.
Xu \textit{et al.}~\cite{xu2020neural} provably show that fully-connected neural networks can extrapolate well when the task is linear and the geometry of the training distribution is sufficiently diverse.
Wu \textit{et al.}~\cite{wu2022extrapolation} exhibits that neural networks with Hadamard product (NNs-hp) can learn high degree nonlinear function.
But the general situation is still unknown.
There is ongoing research to improve the robustness of machine learning models to distribution shifts. 
Approaches to addressing OOD generalization include training on diverse data and multiple tasks~\cite{albuquerque2020improving}, using adversarial training to enhance the model's ability to recognize OOD data~\cite{yi2021improved, ryu-etal-2018-domain}, and incorporating domain-specific knowledge into the model~\cite{wang2022out}.

Weight averaging is a method that has recently been shown to greatly improve the out-of-distribution generalization ability of machine learning models~\cite{izmailov2018averaging, cha2021swad, rame2022diverse, wortsman2022model}. 
It involves averaging the weights of different models and is similar to ensemble learning~\cite{sagi2018ensemble}, although the two approaches are essentially different, as we will discuss in Section~\ref{sec:thesis_related}. 
There are two types of weight averaging: averaging multiple checkpoints along the training trajectory of a single model~\cite{izmailov2018averaging, cha2021swad}, and averaging the weights of multiple individually trained models~\cite{rame2022diverse, wortsman2022model}. 
The latter approach has been found to be more effective and will be the focus of our work.

\begin{figure}[h]
    \centering
    \includegraphics[width=0.7\textwidth]{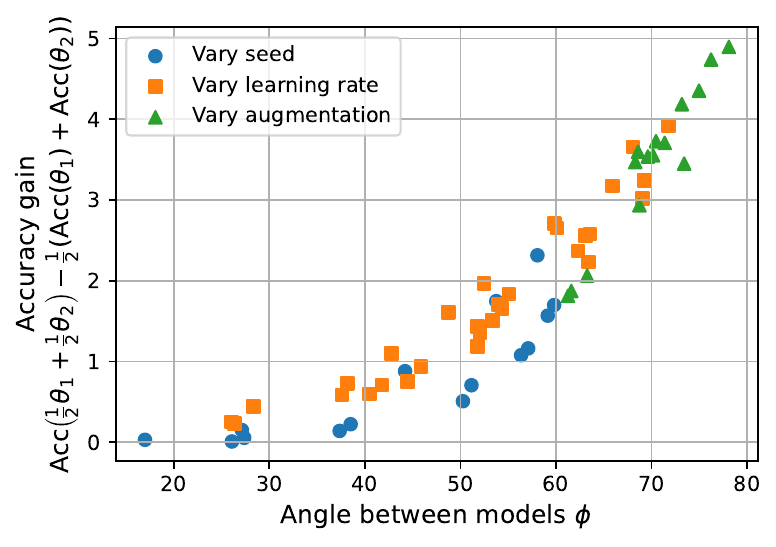}
    \caption{Figure originally published in \textit{Model Soups} \cite{wortsman2022model}. Each dot represents a pair of models trained with different hyperparameters. Larger difference between the hyperparameters corresponds to larger model angle $\phi$.}
    \label{fig:angles}
\end{figure}

Figure~\ref{fig:angles}, adapted from \textit{Model Soups}~\cite{wortsman2022model}, illustrates an interesting phenomenon: there is a strong correlation between the accuracy gain of a weight averaged model and the angles between the models. 
As shown in the plot, larger angles between two models result in greater accuracy gain for the averaged model. 
This suggests that increasing model diversity can lead to better performance for the averaged model. However, the authors of \cite{wortsman2022model} do not explicitly take advantage of this phenomenon in their work.

Inspired by the previous works and this interesting phenomenon, we propose to explicitly increasing model diversity by incorporating gradient diversity into weight averaging.
Specifically, we train a set of individual models on the same in-distribution data and add a regularization term to the classification loss that measures the gradient similarity between pairs of models.
This regularization term serves to enforce gradient diversity between the models during training, thereby increasing model diversity. 
The final model is obtained by averaging the weights of the trained models, which is then tested on out-of-distribution data.

Another focus in this thesis is few-shot domain adaptation.
Domain adaptation is a machine learning problem that involves adapting a model trained on one domain, or distribution of data, to perform well on a different but related domain.
There are several approaches to addressing the problem of domain adaptation, including feature-based methods~\cite{li2018domain}, which aim to find a common feature space that is invariant across domains, and parameter-based methods~\cite{tzeng2017adversarial}, which aim to learn domain-invariant model parameters. 
Domain adaptation can also be viewed as a transfer learning problem~\cite{long2015learning, long2017deep}, in which the goal is to transfer knowledge from a source domain, where the model has been trained, to a target domain, where the model will be used.

In contrast to out-of-domain (OOD) generalization, where no target data is available, domain adaptation is a type of machine learning that allows access to target data during training.
There are three categories of domain adaptation based on the amount and type of accessible target data: (i) unsupervised domain adaptation~\cite{ganin2015unsupervised, saito2018maximum}, where only unlabeled target data is available; (ii) semi-supervised domain adaptation~\cite{li2021cross, saito2019semi}, where both labeled and unlabeled target data are available, but the labeled data is limited; (iii) supervised domain adaptation, where only labeled target data is available.
Few-shot domain adaptation is a special case of supervised domain adaptation, where only a few target data points are available~\cite{li2021few, teshima2020few}.

Based on the observation that weight averaging can improve the performance for OOD generalization greatly, we aim to investigate the use of weight averaging to improve the performance of few-shot domain adaptation. 
We adopt two approaches: (i) adapting each model trained on the source domain individually with few-shot samples, averaging the models, and testing the averaged model on the target domain; and (ii) weight averaging the models first, then adapting the averaged model with few-shot smaples, and testing it on the target domain. 
Our goal is to determine the extent to which weight averaging can improve performance in few-shot domain adaptation, without using gradient diversity.

In the experiment part, we perform a comprehensive experimental study.
For the main results, we use weight averaging with gradient diversity in both in-distribution setting, where CIFAR100~\cite{krizhevsky2009learning} is used, and out-of-distribution setting, where datasets from DomainBed~\cite{gulrajani2020search} are used.
We find that weight averaging can not only increase the performance for OOD generalization but also improve the in-distribution performance.
For few-shot domain adaptation, we mainly conduct experiments on digits datasets and VisDA-C~\cite{visda2017}.
The experiment results show that weight averaging can improve the few-shot domain adaptation accuracy significantly.

\textbf{Paper outline.} Section~\ref{sec:thesis_related} gives a review of previous related works. Section~\ref{sec:thesis_background} presents the background knowledge about this paper for comprehension purpose. Our proposed method and experiments are covered in Section~\ref{sec:thesis_method} and Section~\ref{sec:thesis_experiments} respectively. Section~\ref{sec:thesis_discussion} concludes this paper and details the scope and limitations of our method.\clearpage
\section{Related Works}
\label{sec:thesis_related}
In this section, we review relevant research on out-of-distribution generalization and few-shot domain adaptation.
We begin by discussing the general introduction of OOD generalization and domain adaptation, followed by some representative methods and techniques that are applied to solve these issues.
We then provide an overview of weight averaging and sharpness-aware minimization before proceeding to our own contributions.

\subsection{Out-of-Distribution Generalization}

Developing machine learning models that are robust to distribution shift and able to generalize well to out-of-distribution scenarios is a critical challenge in practical applications. 
To address this problem, significant efforts have been made to training models that can effectively perform in these scenarios~\cite{barbu2019objectnet, beede2020human, hendrycks2019benchmarking, michaelis2019benchmarking, schott2021visual, wenzel2022assaying}.

In the study conducted by Schott \textit{et al.}~\cite{schott2021visual}, the authors posited that the generation process for the training images is controlled by certain "Factor of Variations" (FoVs). They then sought to evaluate whether models could learn and predict these FoVs on test splits. However, they found that none of the models were able to effectively generalize to out-of-distribution (OOD) scenarios, in which the FoVs were not present in the training data. There are several potential explanations for this generalization failure, including the use of "shortcut learning" by the models, which may lead them to rely on auxiliary information and spurious correlations rather than true mechanistic relationships \cite{kilbertus2018generalization, ilyas2019adversarial, scholkopf2021toward, geirhos2020shortcut}. The essence is the trained models are prone to use the auxiliary information and the spurious correlations \cite{kirichenko2022last}, rather than the true mechanistic relationships. 
Another issue is the inconsistency of distributions, as models may experience significant performance degradation when tested on data that is not drawn from the same distribution as the training data.

A variety of approaches have been proposed to address the problem of out-of-distribution generalization, including unsupervised representation learning~\cite{ruan2021optimal, kim2018disentangling} and supervised learning~\cite{ganin2016domain, motiian2017unified}.
Among these, we focus on a particular optimization algorithm, the sharpness-aware minimization ~\cite{foret2020sharpness, kwon2021asam, andriushchenko2022towards}, which will be discussed in more detail in Section~\ref{sec:sharpness}.

\textbf{Unsupervised Representation Learning}. 
Unsupervised representation learning is training a model to extract and learn meaningful features from input data without the use of labeled examples. 
It has the potential to learn more general and transferable features compared to supervised learning, which relies on labeled data for training.
This is typically done through the use of techniques such as autoencoders, generative models, and clustering algorithms.
\cite{ruan2021optimal} proposed a method to learn optimal representations under covariate shift based on domain-agnostic augmentation and contrastive self-supervised learning (SSL) \cite{wang2021multimodal, radford2021learning, he2020momentum}.
FactorVAE~\cite{kim2018disentangling} improves the $\beta$-VAE~\cite{higgins2017betavae}, trying to extract the latent factors (basic visual concepts) that can be disentangled from image generation process and potentially benefit the OOD generalization.

\textbf{Supervised Methods}. 
Supervised learning approaches for out-of-distribution generalization often focus on learning invariant features from labeled data. Previous research has demonstrated that invariant features learned from different domains can be transferable and robust to domain shift~\cite{muandet2013domain, albuquerque2020adversarial}.
\cite{ganin2016domain} proposes domain-adversarial neural network (DANN) to learn discriminative and domain-invariant features. 
It does this by jointly optimizing three components: the underlying features, the label predictor and the domain classifier.
The goal is to learn the representations in such a way that they are able to ``confuse'' the domain classifier, so that these features are domain invariant.
\cite{motiian2017unified} utilizes the Siamese architecture and semantic feature alignment to learn invariant features. 
This method minimizes the distance between samples from the same class but different domains, and maximize the distance between samples from different domains and different classes.

\subsection{Domain Adaptation}
Domain adaptation is a technique used to address the discrepancy that can occur when applying a trained model to a new domain. Without domain adaptation, the model may perform poorly due to the difference in distribution between the training and test domains. To mitigate this issue, domain adaptation aims to align the source domain (where the model was trained) with the target domain (where the model is being applied).
Like OOD generalization, the methods for domain adaptation can also be categorized into supervised and unsupervised approaches.
For the supervised case, the few-shot domain adaptation receives much more attention.

\textbf{Unsupervised Domain Adaptation.}
Unsupervised methods for domain adaptation offer a solution to the problem of adapting models to a new domain without the need for labeled data from that domain.
A notable example of this is the work by Ganin \textit{et al.}~\cite{ganin2015unsupervised}.
In this work, the authors propose the use of a domain classifier connected to the feature extractor through a gradient reversal layer, which multiplies the gradient by a negative constant during backpropagation-based training. 
This allows the model to effectively adapt to the new domain using only unlabeled data.
\cite{saito2018maximum} propose a method for unsupervised domain adaptation that maximizes the discrepancy between the outputs of two classifiers to identify target samples that are far from the support of the source distribution and align the distributions of the source and target domains.

\textbf{Few-shot Domain Adaptation.}
By leveraging the knowledge learned from the source domain, few-shot domain adaptation can effectively adapt a model to the target domain using only a small amount of labeled data, which makes it more attractive than unsupervised methods when the target data is scarce.
\cite{motiian2017few} uses adversarial learning to minimize the discrepancy between two domains. Because there is a limited amount of target data available, they augment the traditional binary adversarial discriminator to distinguish between four classes by pairing domain labels as well as class labels.
\cite{xu2019d} adopts stochastic neighborhood embedding techniques (SNE) and a modified Hausdorff distance to minimize the distance between samples from the source domain and target domain while maximizing the margin between inter-class distances and minimizing intra-class distances in both domains in order to achieve domain-invariance.

\subsection{Pre-training and Fine-tuning}
Large-scaled pre-training and fine-tuning has achieved great progress in computer vision and natural language processing \cite{yalniz2019billion, kolesnikov2020big, bommasani2021opportunities}. It has been shown that these models can learn representations that transfer well for various downstream tasks \cite{radford2021learning, jia2021scaling, yu2022coca}.
They have become common practices for OOD generalization and domain adaptation.

In \cite{dittadi2021role}, the authors investigate the influence of using pre-trained representations on the ability of reinforcement learning agents to generalize to new, unseen situations.
They found that agents benefit from the pre-training and some properties of the pretrained representations are useful to predict which agents will exhibit superior generalization capabilities.
According to Kirichenko \textit{et al.}~\cite{kirichenko2022last}, it was found that the generalization performance of pre-trained models can be improved by simply fine-tuning the last layer, which is often the final classifier layer, on an un-spurious dataset. This helps to mitigate the effects of spurious correlations that are present in pre-trained models. 
Kumar \textit{et al.}~\cite{kumar2022fine} similarly discovered that fine-tuning the entire network is beneficial for in-distribution data but not for out-of-distribution (OOD) data. They recommend that practitioners first conduct linear-probing on the pre-trained model and then fine-tune the model specifically for OOD tasks.

Pre-training and fine-tuning are also used in the weight averaging method.
Researchers emphasize that in order for this method to be effective, all individual models must be initialized using the same pre-trained weights, as previous research~\cite{frankle2020linear, neyshabur2020being} has shown that linear averaging of weights from a shared initialization can lead to small loss values.

\subsection{Ensemble Learning}
Ensemble learning is a long-researched method of combining multiple models to improve the overall performance of the system.
The basic idea is that by combining the predictions of multiple models, the ensemble will reduce errors and increase the robustness of the predictions. 
There are several different techniques for ensemble learning, including bagging~\cite{breiman1996bagging}, boosting~\cite{bauer1999empirical, freund1997decision} and random forest~\cite{breiman2001random}.

In recent years, many works have shown that ensemble can not only improve the final performance for in-distribution data, but also exhibit high accuracy for OOD scenarios \cite{ovadia2019can, mustafa2020deep}. 
In \cite{ross2020ensembles}, the authors present a new method for measuring the diversity of two models' generalization capabilities over small patches of the data manifold.
By enforcing the diversities in ensemble learning, they improved the final performance under covariate shift. 
\cite{teney2022evading} propose a similar idea by using the gradient similarity as the regularizer in ensemble training, which mitigates the simplicity bias \cite{hermann2020shapes, neyshabur2014search, shah2020pitfalls} and improves the OOD generalization.

Ensemble learning is different from weight averaging.
Instead of averaging the model predictions, WA averages the model weights.

\subsection{Flatness and Generalization}
\label{sec:sharpness}
The concept of flatness in the context of machine learning and optimization can be traced back to 1994~\cite{hochreiter1994simplifying}.
In recent years, many studies have highlighted the relationship between the flatness of minima in the loss landscape and generalization performance~\cite{keskar2016large, neyshabur2017exploring, dinh2017sharp}.

Contrary to traditional empirical risk minimization (ERM) that directly minimize the loss functions, \cite{foret2020sharpness} proposed a method called \textbf{Sharpness-Aware Minimization (SAM)} that can simultaneous minimize the loss value and loss sharpness.
It aims to find model parameters that are located in a flat region of the loss landscape where nearby points also have low loss values.

Another line of works that can be used to achieve flat minima is weight averaging. 
\textbf{Stochastic Weight Averaging (SWA)}~\cite{izmailov2018averaging} averages multiple checkpoints along the training trajectory of a single model.
With the use of a cyclical or constant learning rate, it has been shown that SWA can find flatter solutions than stochastic gradient descent, resulting in better generalization.

\subsection{Weight Averaging}
As we have mentioned above, one of the advantages of weight averaging is it can lead to flatter minima.

Vanilla SWA has some drawbacks like too few weights and inaccurate approximation of the flat minima. To overcome these problmes, \cite{cha2021swad} improve SWA by utilizing a dense and overfit-aware sampling strategy, named Stochastic Weight Averaging Densely (SWAD), to gather the stochastic weights along the training trajectory.
Apart form SWAD, periodic SWA (PSWA)~\cite{guo2022stochastic} empirically evaluates the effect of cyclical or high constant learning rate. \cite{gupta2020stochastic} propose a paralleled version of SWA (SWAP) that improves the generalization performance.
Besides, there are also some works explore to combine ensemble and weight averaging. \cite{arpit2021ensemble} propose a method to ensemble the moving averaged models, which results in further performance improvement.

Recently, there have been some studies on weight averaging that differ from averaging checkpoints from a single training trajectory.
DiWA~\cite{rame2022diverse} and Model Soups~\cite{wortsman2022model} are two concurrent works that average weights from multiple, individually trained model.
Specifically, these methods train several individual models that start from a shared initialization  but with different hyperparameters, and then average their weights. 
This method has been shown to be more effective than averaging weights over the training trajectory of a single model.

While weight averaging and sharpness-aware minimization have been successful in improving OOD generalization, there are still some limitations.
In this paper, we aim to improve the weight averaging method by incorporating gradient similarity as a regularizer to explicitly increase model diversity.
Furthermore, we propose a combination of weight averaging and SAM to address the few-shot domain adaptation problem, which was not adequately explored.

\clearpage
\section{Background}
\label{sec:thesis_background}
In this section, for comprehensiveness and better understanding, we present some necessary background knowledge related to this paper.

\subsection{Learning Paradigms}
There are many ways to categorize the learning paradigms.
Traditionally, people consider three types of learning paradigms: supervised learning, unsupervised learning and reinforcement learning.
But depending on the type of data and the source of the data, there can be a more detailed classification.
In Table~\ref{table:paradigms}, we list some closely related learning setups. In this paper, we only care about the last two types: domain generalization and supervised (few-shot) domain adaptation.

\begin{table}[h]
    \caption{Learning setups.
    $L^d$ and $U^d$ denote the labeled and unlabeled distributions from domain $d$.}
    \begin{center}
    \begin{tabular}{lll}
        \toprule
        \textbf{Setup} & \textbf{Training inputs} & \textbf{Test inputs} \\
        \hline
        Generative learning        & $U^1$ & $\emptyset$ \\
        Unsupervised learning      & $U^1$ & $U^1$ \\
        Supervised learning        & $L^1$ & $U^1$ \\
        Semi-supervised learning   & $L^1, U^1$ & $U^1$ \\
        Multitask learning         & $L^1, \ldots, L^\dtr$ & $U^1, \ldots, U^\dtr$ \\
        Unsupervised domain adaptation    & $L^1, \ldots, L^\dtr, U^{\dtr + 1}$ & $U^{\dtr + 1}$ \\
        \textbf{Supervised domain adaptation}    & $L^1, \ldots, L^\dtr, L^{\dtr + 1}$ & $U^{\dtr + 1}$ \\
        \textbf{Domain generalization}      & $L^1, \ldots, L^\dtr$ & $U^{\dtr + 1}$\\
        \hline
    \end{tabular}
    \end{center}
    \label{table:paradigms}
\end{table}

Few-shot domain adaptation is a special case of supervised domain adaptation, where for the training data $L^{\dtr + 1}$, only a few samples for each class are available.

\subsection{Distribution Shift and Generalization}

Training a model that is robust and generalize well under distribution shifts is a critical problem in real practice.
We first give a formal definition of this problem, and then list some variations.

\textbf{Problem setup}. Given some i.i.d. training data $\{(x_i, y_i) \}_{i=1}^{n_{tr}} \sim p_{tr}(x,y)$, we want to train a predictor $f: x \rightarrow y$ that works well for the test domain:
\begin{equation}
\centering
    \min_f R(f) = \mathbb{E}_{p_{te}} [ \ell (f(x), y) ],
\end{equation}
where $R(f)$ is the risk function or expected loss, $\ell$ is a loss function, which can be the cross-entropy loss function for classification.

Note that this is different from the normal machine learning setting.
Usually, we assume the test data has the same distribution as the training data, i.e., $p_{tr}(x,y) = p_{te}(x,y) = p$, hence the objective is to minimize the loss on the training data:

\begin{equation}
\centering
    \min_f R(f) = \mathbb{E}_{p_{tr}} [ \ell (f(x), y) ],
\end{equation}
In practice, people usually replace the expected loss with empirical loss:

\begin{equation}
\centering
    \min_f \hat R(f) = \frac{1}{n_{tr}} \sum_{i=1}^{n_{tr}} \ell (f(x_i), y_i)
\end{equation}
And the objective of empirical risk minimization (ERM) is to find a predictor $\hat f$ that minimizes this empirical risk:

\begin{equation}
\centering
    \hat f = \arg \min_f \hat R(f)
\end{equation}

However, when $p_{tr}(x,y) \ne p_{te}(x,y) $ there will be a big challenge.
ERM is notorious that the resulted optimized neural network memorizes the training data instead of generalize from it.
Also, the prediction results of the nueral network trained with ERM change drastically when the test sample is out of the training distribution \cite{zhang2017mixup, zhang2021understanding, szegedy2013intriguing}.

\textbf{Various scenarios}. For the shift of the distributions, there are some different variations \footnote{EPFL CIS – RIKEN AIP Seminar: ``Robust machine learning for reliable deployment'', Prof. Masashi Sugiyama, \href{https://www.youtube.com/watch?v=FN7ZPzI63wE}{https://www.youtube.com/watch?v=FN7ZPzI63wE}}.
\begin{itemize}
    \item Full-distribution shift.
    \begin{equation}
        \centering
        p_{tr}(\mathbf x, y) \ne p_{te}(\mathbf x, y)
    \end{equation}
    \item Covariate shift.
    \begin{equation}
        \centering
        p_{tr}(\mathbf x) \ne p_{te}(\mathbf x)
    \end{equation}
    \item Class-prior/target shift.
    \begin{equation}
        \centering
        p_{tr}(y) \ne p_{te}(y)
    \end{equation}
    \item Output noise.
    \begin{equation}
        \centering
        p_{tr}(y|\mathbf x) \ne p_{te}(y|\mathbf x)
    \end{equation}
    \item Class-condition shift.
    \begin{equation}
        \centering
        p_{tr}(\mathbf x|y) \ne p_{te}(\mathbf x|y)
    \end{equation}
\end{itemize}

In this thesis, for the OOD generalization and domain adaptation, we focus on the covraiate shift, where only the distribution of the input $\mathbf{x}$ changes, but the prediction targets $y$ remains unchanged: $p_{tr}(y|\mathbf{x}) = p_{te}(y|\mathbf{x}) = p(y|\mathbf{x})$.
In other words, for classification tasks, the classes are not changed under distribution shifts.

\subsection{Out-of-Distribution Generalization}
Under the covariate shift setting, the formal definition of OOD generalization is as follows:

Let $p_{tr}(x, y) $ be the distribution of the training data, and $p_{te}(x,y)$ be the distribution of the test data.
The posterior distributions of the classes are the same: $p_{tr}(y|x) = p_{te}(y|x)$, but the marginal distributions of the inputs are different:  $p_{tr}(x)\ne p_{te}(x)$.
A machine learning model $f(\cdot, \theta)$ is trained on the training data that minimizes the empirical risk of the training data:
\begin{equation}
    \centering
    \hat{f} (\cdot, \theta) = \arg \min_\theta \frac{1}{N_{tr}} \sum_{i=1}^{N_{tr}} \ell(f(x_i, \theta), y_i), \text{ where } (x_i, y_i) \sim p_{tr}(x,y)
\end{equation}
and also minimizes the empirical risk of the test data at the same time:
\begin{equation}
    \centering
    R(\hat{f}) = \min_f \frac{1}{N_{te}} \sum_{i=1}^{N_{te}} \ell (f(x_i), y_i), \text{ where } (x_i, y_i) \sim p_{te}(x,y)
\end{equation}

\subsection{Few-shot Domain Adaptation}
Different from OOD generalization, where the target data is inaccessible during training, in domain adaptation, the training process can utilize the target data.
But for few-shot domain adaptation, only a few samples for each class of the target data is available. 
There are many ways to do few-shot domain adaptation. For example, adversarial base method~\cite{motiian2017few} and feature based~\cite{xu2019d}.
In this paper, we do few-shot domain adaptation based on fine-tuning.
The formal definition is as follows:

Consider a $C-$class classification problem. Given some i.i.d. samples from the source domain $\{ (x_i, y_i)\}_{i=1}^{N} \sim p_{s}(x,y)$, where $p_s(x,y)$ is the distribution of the source domain data.
We train a neural network $f(\cdot, \theta)$ with the source data to minimize the empirical risk:

\begin{equation}
    \centering
    f (\cdot, \theta) = \arg \min_\theta \frac{1}{N} \sum_{i=1}^{N} \ell(f(x_i, \theta), y_i)
\end{equation}

In addition, we are also given some data from the target domain.
For each class $c \in \{1, 2, \dots, C\}$, there are $k$ target samples: $\{(x_i, y_i)\}_{i=1}^k \sim p_t(x, y) $, where $y_i = c$.
This is called $k$-shot domain adaptation.
Based on the well-trained model $f(\cdot, \theta)$, we fine-tune this model on the $C \times k$ target samples to minimize the empirical loss, and get the model $\hat{f}(\cdot, \theta)$:

\begin{equation}
    \centering
    \hat f (\cdot, \theta) = \arg \min_\theta \frac{1}{C\times k} \sum_{c=1}^{C} \sum_{i=1}^k \ell(f(x_{c,i}, \theta), y_{c,i})
\end{equation}
such that the model $\hat{f}(\cdot, \theta)$ can obtain a small empirical risk on the target domain.

\subsection{Weight Averaging}
The focus of this paper is weight averaging.
Literally, it means averaging the weights of multiple models.
We give the formal definition of weight averaging below.

Given $M$ individual models $\{f(\cdot, \theta_m)\}_{m=1}^M$ that is well trained under some extra conditions, which we will talk about in details in Section~\ref{sec:thesis_method}, the weights of these models can be averaged as follows:

\begin{equation}
    \centering
    f_{\text{WA}} := f(\cdot, \theta_{\text{WA}}), \text{ where } \theta_{\text{WA}} := \frac{1}{M} \sum_{m=1}^M \theta_m
\end{equation}

The final weight averaged model $f_{WA}$ will be used to test on the OOD data directly for OOD generalization or fine-tuned on the target data first for domain adaptation.

\subsection{Sharpness-Aware Minimization}
Sharpness-aware minimization (SAM) is the main optimization algorithm that we are going to use for the project.
For comprehensiveness and comparison, we give the formal definitions for both stochastic gradient descent and sharpness-aware minimization.
More details can be found in \cite{foret2020sharpness, kaddour2022questions}

\textbf{Stochastic Gradient Descent (SGD)}
SGD is the most common optimization algorithm that people use to optimize machine learning models.
In this framework, the objective is a loss function $\mathcal{L}(\theta)$ parameterized by $\theta$.

\begin{equation}
    \centering
    \mathcal{L} (\theta) = \frac{1}{N} \ell (f(x_i, \theta), y_i)
\end{equation}

And the parameters are optimized via a simple update rule, using the negative gradient of the loss function, w.r.t. $\theta$:
\begin{equation}
    \centering
    \theta_{t+1} = \theta_t - \lambda \nabla_\theta \mathcal{L}(\theta)
\end{equation}
where the $\lambda$ is the learning rate. Normally, this algorithm is implemented in a batch manner.

\textbf{Sharpness-Aware Minimization (SAM)}
Intuitively, SAM tries to find an optimum that lies in a flat region of the loss landscape, where its neighbors also have low loss values.
SAM optimizes a loss function in two steps:
first in a given neighborhood $\rho$, it computes the worst-case loss value under $\epsilon$ perturbation; second, it minimize the loss value w.r.t. $\theta$ under the worst-case perturbation.

\begin{equation}
    \centering
    \min_\theta \max_{\| \epsilon \|_2 \le \rho} \mathcal{L} (\theta + \epsilon)
    \label{eq:sam}
\end{equation}
where $\rho \le 0$ is a hyperparameter that controls the size of the neighborhood.

However, directly optimizing Eq.~\ref{eq:sam} is intractable.
Foret \textit{et al.}~\cite{foret2020sharpness} propose to approximate the inner maximization with a first-order Taylor expansion w.r.t. $\theta$ around 0, and hence obtain:

\begin{equation}
    \centering
    \epsilon^*(\theta) = \arg \max_{\| \epsilon \|_2 \le \rho} \mathcal{L}(\theta + \epsilon) \approx \arg \max_{\| \epsilon \|_2 \le \rho} \mathcal{L}(\theta) + \epsilon^T\nabla_\theta \mathcal{L}(\theta) = \arg \max_{\| \epsilon \|_2 \le \rho} \epsilon^T\nabla_\theta \mathcal{L}(\theta) 
\end{equation}

Using dual norm, the solution $\hat \epsilon(\theta)$ is given by:

\begin{equation}
    \centering
    \hat{\epsilon}(\theta) = \rho \cdot \frac{\nabla_\theta \mathcal{L(\theta)}}{\| \nabla_\theta \mathcal{L(\theta)} \| }
\end{equation}

Substituting the $\hat{\epsilon}(\theta)$ back into Eq.~\ref{eq:sam}, we get the final update rule for SAM:
\begin{equation}
    \centering
    \theta_{t+1} = \theta_t - \lambda \nabla_\theta (\max_{\| \epsilon \|_2 \le \rho} \mathcal{L} (\theta + \epsilon) ) \approx \theta_t - \nabla_\theta \mathcal{L(\theta)}|_{\theta + \hat{\epsilon}}
\end{equation}\clearpage
\section{Method}
\label{sec:thesis_method}
In this section, we will present the details of our method for OOD generalization and few-shot domain adaptation.

\subsection{Weight Averaging}
Our method is based on weight averaging. 
The successful application of weight averaging strategy requires some additional conditions to be met.
As pointed out in the work of Alexandre Ramé \textit{et al.}~\cite{rame2022diverse}, shared initialization and mild hyperparameter search are critical requirements for weight averaging.

\textbf{Shared initialization.} All the individual models used for averaging must start from a shared initialization. 
Only in this case can the linear averaging of different models retain a low loss value.
For OOD generalization, we utilize the ImageNet~\cite{deng2009imagenet} pre-trained ResNet50~\cite{he2016deep}, and linear probe the final classification layer on the in-distribution data first to obtain the shared initialization.
When using the small digits datasets in few-shot domain adaptation, we also explore the simple convolutional neural networks (CNN), where the pretrained weights are unavailable.
In this case, we train the simple CNN on the in-domain data to achieve at least 85\% accuracy, and then use this model as shared initialization.

\textbf{Mild hyperparameter search.}
All the individual models used for averaging should be trained with different hyperparameters.
But the selection of hyperparameters is critical because the difference between these hyperparameters can not be too large.
Otherwise, the pretrained weight in the shared initialization might be destroyed and averaging these models directly will result in poor minima.

In Table~\ref{tab:hyper} we list the default values of the hyperparameters and their distributions for random search.

\begin{table}[ht]
    \centering
    \begin{tabular}{c|c|c}
        \hline
        \textbf{Hyperparameter} & \textbf{Default value} & \textbf{Random distribution}  \\
        \hline
        Learning rate  & $5\cdot 10^{-5}$ & [1, 3, 5]$\cdot 10^{-5}$\\
        Weight decay   & 0    & [$10^{-4}, 10^{-6}$] \\
        SAM $\rho$     & 0.05 & [0.01, 0.02, 0.05, 0.1] \\
        ResNet dropout & 0    & [0, 0.1l 0.5] \\
        \hline
    \end{tabular}
    \caption{Default values of the hyperparameters and their distributions for random search.}
    \label{tab:hyper}
\end{table}


\subsection{OOD Generalization}
We extend the weight averaging by 
adding the gradient similarity as a regularizer in order to explicitly increase the resulted model diversity.
For out-of-distribution generalization, the pipeline is composed of 3 steps:

\textbf{Step 1: Initialization} As we have mentioned above, we linear probe the last \texttt{fc} layer of an ImageNet pretrained ResNet50 to get the shared initialization.
In this process, we use the default values for the hyperparameters.

\textbf{Step 2: Sweep training} Using the shared initialization in step 1, we will launch multiple runs to fine-tune the model with different hyperparameters. 
The hyperparameter values are randomly selected and combined from their corresponding random distribution, as listed in Table~\ref{tab:hyper}.

\textbf{Step 3: Weight averaging} After sweep training, we will average all the model weights to obtain the final model.
In this thesis, we use the uniform selection which is uniformly averaging all the models.
But in previous works~\cite{rame2022diverse, wortsman2022model}, the researchers also explored constrained/greedy selection, where only the models that will improve the final performance will be averaged.

\textbf{Gradient Similarity.} Suppose in sweep training, we are going to launch $k$ individual runs.
For each run, two separate models will be trained. Hence in total, it will result in $2k$ models.
The gradient similarity is computed between the pair of models in each run.
The detailed illustration of weight averaging training with gradient similarity is shown in Algorithm~\ref{alg:wa_grad}.

\begin{algorithm}[ht]
\caption{Weight Averaging with Gradient Diversity}
\label{alg:wa_grad}
\begin{algorithmic}
\Require ImageNet pretrained \texttt{ResNet50} $f$, in-domain data $\mathcal{D}_{in}$, out-of-domain data $\mathcal{D}_{out}$

\vspace{0.3cm}

\Procedure{\texttt{fine\_tune}}{$f_{\theta_{k1}}, f_{\theta_{k2}}, \mathcal{D}$} 
\ForAll{sample $(x, y) \in \mathcal{D}$ }
    \State $loss_{k1} \gets \ell(f_{\theta_{k1}}(x), y) + cossim(\nabla_{h_{k1}} f_{\theta_{k1}}(x), \nabla_{h_{k2}} f_{\theta_{k2}}(x)))$
    \State $loss_{k2} \gets \ell(f_{\theta_{k2}}(x), y) + cossim(\nabla_{h_{k1}} f_{\theta_{k1}}(x), \nabla_{h_{k2}} f_{\theta_{k2}}(x)))$
    \State $f_{\theta_{k1}} \gets $ \texttt{UPDATE}($f_{\theta_{k1}}, loss_{k1}$) 
    \State $f_{\theta_{k2}} \gets $ \texttt{UPDATE}($f_{\theta_{k2}}, loss_{k2}$) 
\EndFor
\State \textbf{return} $f_{\theta_{k1}}, f_{\theta_{k2}}$
\EndProcedure

\vspace{0.3cm}

\Procedure{\texttt{Main}}{$f, \mathcal{D}_{in}$}
\State $f_{\theta_{init}} \gets $ \texttt{LINEAR\_PROBE}($f, \mathcal{D}_{in}$)
\State Initialize $2k$ models $f_{\theta_{k1}} = g_{k1} \circ h_{k1}, f_{\theta_{k2}} = g_{k2} \circ h_{k2}$ using the shared initialization $f_{\theta_{init}} $ 

\For{each pair of models $f_{\theta_{k1}}, f_{\theta_{k2}} $ }
    \State $f_{\theta_{k1}}, f_{\theta_{k2}} \gets $ \texttt{FINE\_TUNE}($f_{\theta_{k1}}, f_{\theta_{k2}}, \mathcal{D}_{in}$)
\EndFor

\ForAll{models $f(\cdot, \theta_i)$}
    \State $f(\cdot, \hat{\theta}) \gets f(\cdot, \frac{1}{2k}\sum_{i=1}^{2k}\theta_i)$
\EndFor

\State \textbf{return} $f_{\hat{\theta}}$
\EndProcedure

\end{algorithmic}
\end{algorithm}

\subsection{Few-Shot Domain Adaptation}

For few-shot domain adaptation, apart from the above 3 steps, there is an extra adaptation step.
In this step, we select $k$ samples per class from the target domain, and use them to fine-tune the models.
This is called $k$-shot adaptation.
We also explore two strategies for adaptation: (i) adaptation after weight averaging and (ii) adaptation before weight averaging.

\textbf{Step 4:} Few-shot adaptation after weight averaging and few-shot adaptation before weight averaging are shown in Algorithm~\ref{alg:adapt_after_wa} and Algorithm~\ref{alg:adapt_before_wa} respectively.

\begin{algorithm}[ht]
\caption{Few-Shot Adaptation \textbf{after} Weight Averaging}
\label{alg:adapt_after_wa}
\begin{algorithmic}
\Require $2k$ fine-tuned models, training split of the target domain dataset $\mathcal{D}_{tr}$, test split of the target domain dataset $\mathcal{D}_{te}$, number of adapted samples per class $k$

\vspace{0.3cm}

\Procedure{\texttt{Adapt}}{$f(\cdot, \theta), \mathcal{D}_{tr}, k$} 
    \State $\hat{\mathcal{D}}  \gets $ select $k$ samples per class from $\mathcal{D}_{tr}$
    \State $f(\cdot, \theta) \gets $ \texttt{FINE\_TUNE}($f(\cdot, \theta), \hat{\mathcal{D}}$)
    \State \textbf{return } $f(\cdot, \theta)$
\EndProcedure

\vspace{0.3cm}

\Procedure{Main}{$f(\cdot, \theta_1), \cdots, f(\cdot, \theta_{2k}), \mathcal{D}_{tr}, \mathcal{D}_{te}$}
\ForAll{models $f(\cdot, \theta_i)$}
    \State $f(\cdot, \hat{\theta}) \gets f(\cdot, \frac{1}{2k}\sum_{i=1}^{2k}\theta_i)$
\EndFor
\State $f(\cdot, \hat{\theta}) \gets $ \texttt{ADAPT}($f(\cdot, \hat{\theta}), \mathcal{D}_{tr}, k$)
\State \texttt{acc} $\gets $ \texttt{TEST}($f(\cdot, \hat{\theta}), \mathcal{D}_{te}$)
\State \textbf{return} \texttt{acc}
\EndProcedure

\end{algorithmic}
\end{algorithm}

\begin{algorithm}[ht]
\caption{Few-Shot Adaptation \textbf{before} Weight Averaging}
\label{alg:adapt_before_wa}
\begin{algorithmic}
\Require $2k$ fine-tuned models, training split of the target domain dataset $\mathcal{D}_{tr}$, test split of the target domain dataset $\mathcal{D}_{te}$, number of adapted samples per class $k$

\vspace{0.3cm}

\Procedure{\texttt{Adapt}}{$f(\cdot, \theta), \mathcal{D}_{tr}, k$} 
    \State $\hat{\mathcal{D}}  \gets $ select $k$ samples per class from $\mathcal{D}_{tr}$
    \State $f(\cdot, \theta) \gets $ \texttt{FINE\_TUNE}($f(\cdot, \theta), \hat{\mathcal{D}}$)
    \State \textbf{return } $f(\cdot, \theta)$
\EndProcedure

\vspace{0.3cm}

\Procedure{Main}{$f(\cdot, \theta_1), \cdots, f(\cdot, \theta_{2k}), \mathcal{D}_{tr}, \mathcal{D}_{te}$}
\ForAll{models $f(\cdot, \theta_i)$}
    \State $f(\cdot, \theta_i) \gets$ \texttt{ADAPT}($f(\cdot, \theta_i), \mathcal{D}_{tr}, k$)
\EndFor

\ForAll{models $f(\cdot, \theta_i)$}
    \State $f(\cdot, \hat{\theta}) \gets f(\cdot, \frac{1}{2k}\sum_{i=1}^{2k}\theta_i)$
\EndFor

\State \texttt{acc} $\gets $ \texttt{TEST}($f(\cdot, \hat{\theta}), \mathcal{D}_{te}$)
\State \textbf{return} \texttt{acc}
\EndProcedure

\end{algorithmic}
\end{algorithm}
\clearpage
\section{Experiments}
\label{sec:thesis_experiments}

In this section, we present our key experimental results. These results are divided into two parts: experiments for out-of-distribution generalization and experiments for few-shot domain adaptation. For each part, we begin by describing the experimental setup, followed by a detailed description of our findings. We also include corresponding ablation studies for further analysis.

\subsection{Out-of-Distribution Generalization}
We propose to add model gradient similarity as a regularizer to improve the model diversity in weight averaging.

\subsubsection{Experimental Setup}
In this part, we conduct experiments on CIFAR100~\cite{krizhevsky2009learning} for in-distribution performance verification and on a few datasets from DomainBed~\cite{gulrajani2020search} for out-of-distribution performance evaluation. 

\begin{figure}[ht]
    \centering
    \includegraphics[width=\textwidth]{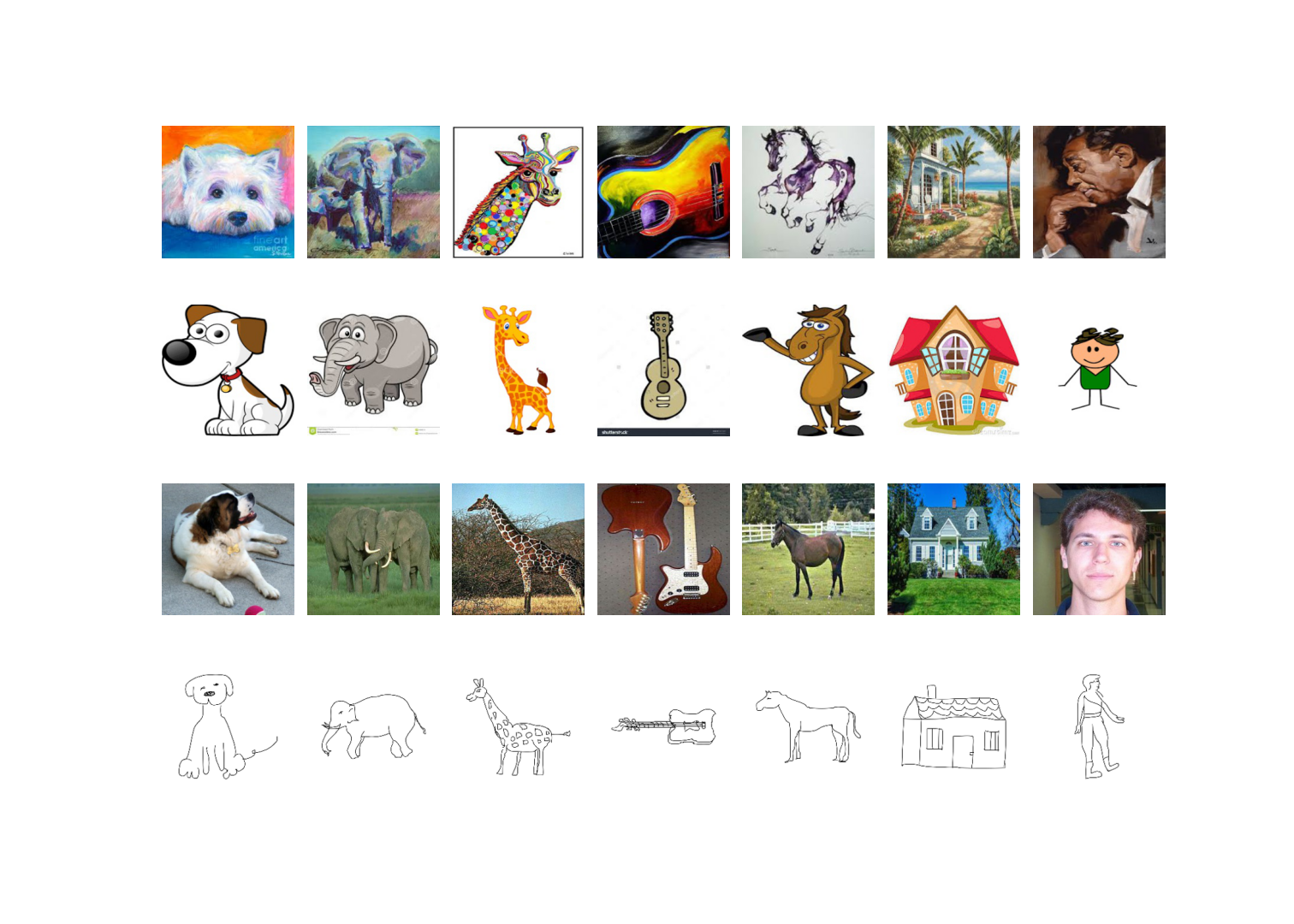}
    \caption{Samples from PACS~\cite{li2017deeper}. From top to bottom are 4 domains: art, cartoon, photo and sketch. From left to right are 7 classes: dog, elephant, giraffe, guitar, horse, house and person.}
    \label{fig:pacs}
\end{figure}

\begin{figure}[ht]
    \centering
    \includegraphics[width=\textwidth]{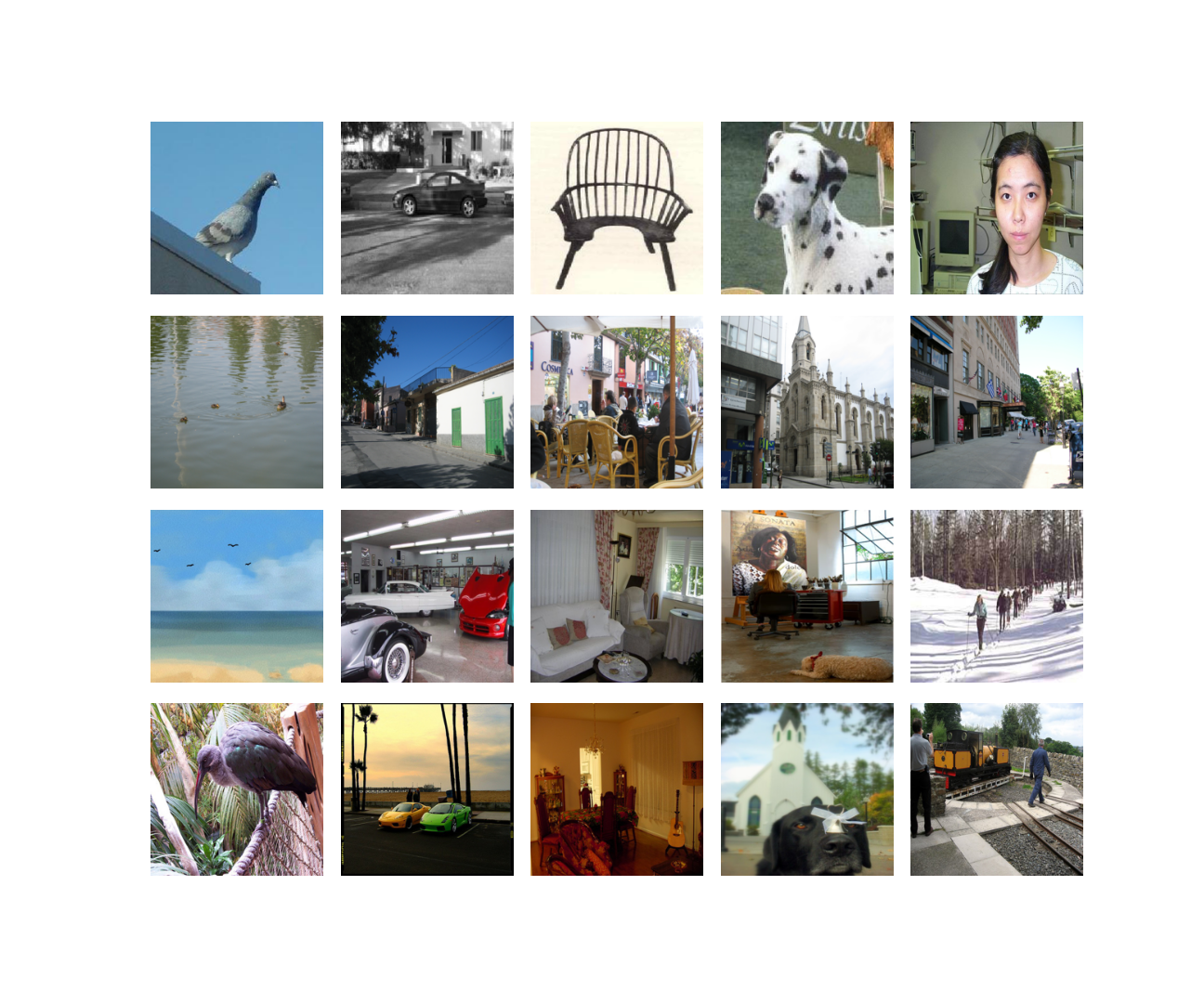}
    \caption{Samples from VLCS~\cite{fang2013unbiased}. From top to bottom are 4 domains: Caltech101, LabelMe, SUN09 and VOC2007. From left to right are 5 classes: bird, car, chair, dog and person.}
    \label{fig:vlcs}
\end{figure}

\textbf{Datasets}. Following previous works~\cite{rame2022diverse, cha2021swad}, we use DomainBed, which is a fair benchmark for evaluating OOD performance.
It contains 5 datasets: PACS~\cite{li2017deeper}, VLCS~\cite{fang2013unbiased}, OfficeHome~\cite{venkateswara2017deep}, TerraIncognita~\cite{beery2018recognition}, and DomainNet~\cite{peng2019moment}.
Due to the time and computation limitations, we conduct experiments on two of them: PACS and VLCS.
PACS dataset contains 9,991 images, 7 classes (dog, elephant, giraffe, guitar, horse, house and person), and 4 domains (\textbf{P}hoto, \textbf{A}rt, \textbf{C}artoon, and \textbf{S}ketch) while VLCS dataset contains 10,729 images, 5 classes (bird, car, chair, dog and person), and 4 domains (Images are collected from 4 datasets: PASCAL \textbf{V}OC~\cite{everingham2010pascal}, \textbf{L}abelMe~\cite{russell2008labelme}, \textbf{C}altech101~\cite{fei2004learning} and \textbf{S}UN09~\cite{choi2010exploiting}).
Some example images from the two datasets are visualized in Fig.~\ref{fig:pacs} and Fig.~\ref{fig:vlcs}.
During training, we resize and crop each image to $224 \times 224$.
Color jittering and random horizontal flips are used for augmentation.

\textbf{Model and Baselines}.
We adopt ResNet50~\cite{he2016deep} as the model architecture.
ERM is the standard Empirical Risk Minimization with Adam optimizer.
DiWA~\cite{rame2022diverse} is the most close work to our paper, but they do not contain the in-distribution results.
We add some experiments on CIFAR100 for in-distribution performance verification.
For a fair comparison, we re-train DiWA on CIFAR100 using the default hyperparameter settings.
For experiments on DomainBed, we compare with Coral~\cite{sun2015return}, SWAD~\cite{cha2021swad}, MA (Moving Average)~\cite{arpit2021ensemble} and DiWA~\cite{rame2022diverse}.
Their results on DomainBed are directly adopt from their original papers.
We use uniform model selection, i.e., averaging all the individual models to obtain the final averaged model.

\textbf{Our Procedure}. The experiments are conducted on each domain separately. For example, for dataset PACS, we consider each domain as the out-of-distribution data (out-domain) successively, and use the rest 3 domains as the in-distribution data (in-domain).
In the first step, we set a shared initialization model using linear probing. The base model is ImageNet pretrained ResNet50.
In the second step, we launch several runs using the shared initialization model, and finetuning them with different hyperparameters individually.
Finally, we average the weights from these models and test it on the out-domain data.

\subsubsection{Results on CIFAR100}
To verify the effectiveness of weight averaging and gradient similarity on in-distribution data, we conduct some experiments on CIFAR100.
Due to time and computation resource consideration, instead of training 60 models (which is the number of models used in \cite{rame2022diverse}), we train 10 individual models.
The experiment results are presented in Table~\ref{tab:cifar100_results}.
From this table, we can see that weight averaging with gradient similarity optimized by SAM achieves the best performance.

\begin{table}[ht]
    \footnotesize
    \centering
    \begin{tabular}{c|c}
        \hline
        \textbf{Setting} & \textbf{CIFAR100}\\
        \hline
        ERM Baseline & 0.7954 \\
        ERM + grad. & 0.8398 \\
        \hline        
        DiWA~\cite{rame2022diverse} & 0.8412 \\
        \hline
        WA (SAM + grad) & \textbf{0.8431} \\
        WA (Adam + grad) & 0.8411 \\
        \hline
    \end{tabular}
    \caption{Experiment results on CIFAR100. ERM Baseline is the standard empirical risk minimization of an ImageNet pretrained ResNet50 on CIFAR100 with Adam optimizer. ERM+grad. is adding the gradient similarity regularizer during ERM training. For our method, we train two models with Adam and SAM optimizer separately.}
    \label{tab:cifar100_results}
\end{table}

\subsubsection{Results on DomainBed}
To evaluate the effectiveness of incorporating gradient similarity into weight averaging, we conducted extensive experiments on the DomainBed dataset. The results of these experiments, which were conducted on the PACS and VLCS datasets, can be seen in Table~\ref{tab:pcas_results} and Table~\ref{tab:vlcs_results}, respectively. On the PACS dataset, our method achieved the best performance on the \texttt{Cartoon} domain and comparable results to DiWA with fewer averaged models. On the VLCS dataset, our method achieved the best performance on the \texttt{Caltech101} domain, but was slightly inferior to the state-of-the-art method.

\begin{table}[ht]
    \footnotesize
    \centering
    \begin{tabular}{c|c|c|c|c|c|c|c}
        \midrule
        \textbf{Method} & \textbf{Weight selection} & \textbf{Init} & \textbf{A} & \textbf{C} & \textbf{P} & \textbf{S} & \textbf{Avg}\\
        \midrule
        ERM             & N/A   & \multirow{4}{*}{Random} & 84.7 $\pm$ 0.4 & 80.8 $\pm$ 0.6 & 97.2 $\pm$ 0.3 & 79.3 $\pm$ 1.0 & 85.5 $\pm$ 0.2 \\
        Coral\cite{sun2015return}   & N/A &                & 88.3 $\pm$ 0.2 & 80.0 $\pm$ 0.5 & 97.5 $\pm$ 0.3 & 78.8 $\pm$ 1.3 & 86.2 $\pm$ 0.3 \\
        SWAD \cite{cha2021swad}  & Overfit-aware   &     & 89.3 $\pm$ 0.5 & 83.4 $\pm$ 0.6 & 97.3 $\pm$ 0.3 & 82.5 $\pm$ 0.8 & 88.1 $\pm$ 0.1 \\
        MA \cite{arpit2021ensemble} & Uniform      &      & 89.1 $\pm$ 0.1 & 82.6 $\pm$ 0.2 & 97.6 $\pm$ 0.0 & 80.5 $\pm$ 0.9 & 87.5 $\pm$ 0.2 \\
        \midrule
        ERM & N/A & \multirow{5}{*}{LP \cite{kumar2022fine}} & 86.8 $\pm$ 0.8 & 80.6 $\pm$ 1.0 & 97.4 $\pm$ 0.4 & 78.7 $\pm$ 2.0 & 85.9 $\pm$ 0.6 \\
        MA \cite{arpit2021ensemble} & Uniform &      & \underline{89.5} $\pm$ 0.1 & 82.8 $\pm$ 0.2 & \underline{97.8} $\pm$ 0.1 & 80.9 $\pm$ 1.3 & 87.8 $\pm$ 0.3 \\
        DiWA~\cite{rame2022diverse}   & Restricted: $M \leq 20$ &           & 89.3 $\pm$ 0.2 & 82.8 $\pm$ 0.2 & 98.0 $\pm$ 0.1 & 82.0 $\pm$ 0.9 & 88.0 $\pm$ 0.3 \\
        DiWA~\cite{rame2022diverse}  & Uniform: $M=60$   &      & \textbf{90.6}  & \underline{83.4} & \textbf{98.2}  & \textbf{83.8}  & \textbf{89.0} \\
        \midrule
        WA+grad & Uniform: \textit{M}=40 & LP~\cite{kumar2022fine} & 89.06 & \textbf{83.49} & 97.54 & \underline{83.74} & \underline{88.46} \\
        \midrule
    \end{tabular}
    \caption{Accuracy on PACS with ResNet50. }
    \label{tab:pcas_results}
\end{table}

\begin{table}[ht]
    \footnotesize
    \centering
    \begin{tabular}{c|c|c|c|c|c|c|c}
        \midrule
        \textbf{Method} & \textbf{Weight selection} & \textbf{Init} & \textbf{C} & \textbf{L} & \textbf{S} & \textbf{V} & \textbf{Avg}\\
        \midrule
    ERM  & N/A & \multirow{4}{*}{Random} & 97.7 $\pm$ 0.4 & 64.3 $\pm$ 0.9 & 73.4 $\pm$ 0.5 & 74.6 $\pm$ 1.3 & 77.5 $\pm$ 0.4 \\
    Coral\cite{sun2015return} & N/A & & 98.3 $\pm$ 0.1 & \textbf{66.1} $\pm$ 1.2 & 73.4 $\pm$ 0.3 & 77.5 $\pm$ 1.2 & \underline{78.8} $\pm$ 0.6 \\
    SWAD \cite{cha2021swad} & Overfit-aware & & 98.8 $\pm$ 0.1 & 63.3 $\pm$ 0.3 & \textbf{75.3} $\pm$ 0.5 & \textbf{79.2} $\pm$ 0.6 & \textbf{79.1} $\pm$ 0.1 \\
    MA \cite{arpit2021ensemble} & Uniform & & \underline{99.0} $\pm$ 0.2 & 63.0 $\pm$ 0.2 & \underline{74.5} $\pm$ 0.3 & 76.4 $\pm$ 1.1 & 78.2 $\pm$ 0.2 \\
        \midrule
    ERM & N/A & \multirow{5}{*}{LP \cite{kumar2022fine}} & 98.1 $\pm$ 0.3 & \underline{64.4} $\pm$ 0.3 & 72.5 $\pm$ 0.5 & 77.7 $\pm$ 1.3 & 78.1 $\pm$ 0.5 \\
    MA \cite{arpit2021ensemble} & Uniform & & 98.9 $\pm$ 0.0 & 62.9 $\pm$ 0.5 & 73.7 $\pm$ 0.3 & 78.7 $\pm$ 0.6 & 78.5 $\pm$ 0.4 \\
    DiWA~\cite{rame2022diverse} & Restricted: $M \leq 20$ & & 98.4 $\pm$ 0.0 & 64.1 $\pm$ 0.2 & 73.3 $\pm$ 0.4 & 78.1 $\pm$ 0.8 & 78.5 $\pm$ 0.1 \\
    DiWA~\cite{rame2022diverse} & Uniform: $M=60$ & & 98.9 & 62.4 & 73.9 & \underline{78.9} & 78.6 \\
    \midrule
    WA+grad & Uniform: \textit{M} = 30 & LP~\cite{kumar2022fine} & \textbf{99.01} & 62.09 & 74.07 & 76.27 & 77.86\\
    \midrule
    
    \end{tabular}
    \caption{Accuracy on VLCS with ResNet50. }
    \label{tab:vlcs_results}
\end{table}

\subsubsection{Ablation Study: The Number of Models for Weight Averaging}
\begin{figure}[ht]
    \centering
    \includegraphics[width=0.7\textwidth]{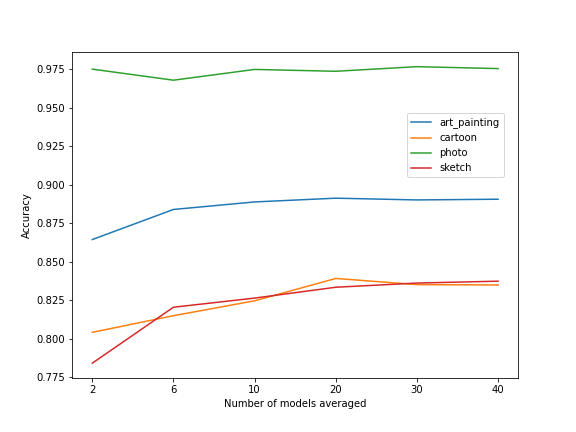}
    \caption{The effect of the number of models used for weight averaging on PACS.}
    \label{fig:pacs_model_num_effect}
\end{figure}

\begin{figure}[ht]
    \centering
    \includegraphics[width=0.7\textwidth]{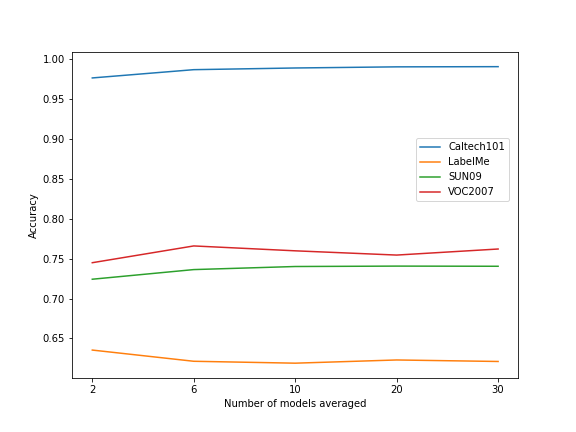}
    \caption{The effect of the number of models used for weight averaging on VLCS.}
    \label{fig:vlcs_model_num_effect}
\end{figure}

In this part, we conduct ablation studies to evaluate the effect of the number of models used for averaging on the PACS and VLCS datasets.
For the PACS dataset, we average 2, 6, 10, 20, 30 and 40 models separately.
Similarly, for the VLCS dataset, we average 2, 6, 10, 20 and 30 models separately.
Given the number of models \texttt{n}, we test the performance three times, where in each trial,
we randomly select \texttt{n} models from all the individual models.
The final performance is the mean value of the three trials.
The results are presented in figures Fig.\ref{fig:pacs_model_num_effect} and Fig.\ref{fig:vlcs_model_num_effect} for PACS and VLCS datasets respectively.

Our analysis of the PACS and VLCS datasets shows that, with the exception of the \texttt{photo} domain for PACS and the \texttt{LabelMe} domain for VLCS, increasing the number of models used for averaging from 2 to 6 greatly improves the final performance. However, when the number of models exceeds 20, the improvement becomes negligible. This observation is similar for both datasets, indicating that a moderate number of models is enough for achieving satisfying performance.

\subsection{Few Shot Domain Adaptation}
In the previous subsection, we have demonstrated the effectiveness of weight averaging for out-of-distribution generalization. 
To further showcase its capabilities, we propose to apply this method to solve the few-shot domain adaptation problem.

\subsubsection{Experimental Setup}
We conduct two sets of experiments on two types of datasets: (i) small-scale digits datasets: MNIST~\cite{lecun1998gradient} , MNIST-M~\cite{ganin2016domain}, USPS~\cite{hull1994database}, SVHN~\cite{netzer2011reading} and
(ii) a large dataset for domain adaptation: VisDA-C dataset~\cite{visda2017}.

\begin{figure}[ht]
    \centering
    \includegraphics[width=\textwidth]{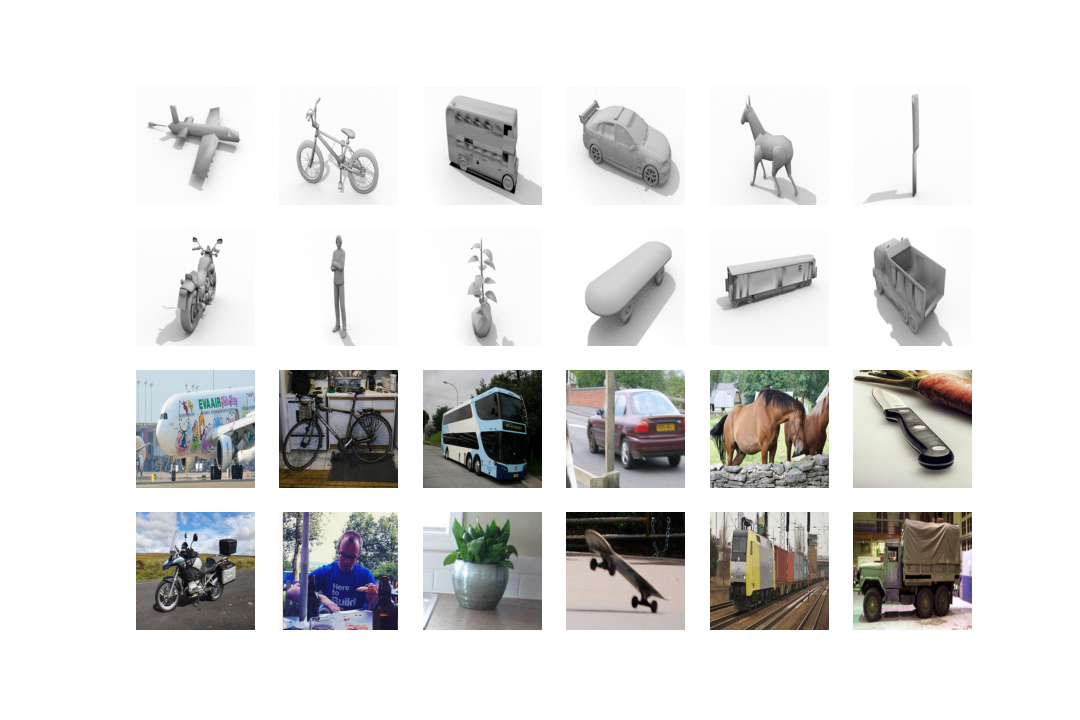}
    \caption{Samples from VisDA-C. There are 12 classes in VisDA-C. From left to right and from up to bottom, they are: aeroplane, bicycle, bus, car, horse, knife, motorcycle, person, plant, skateboard, train and truck. The first 2 lines show the source (train) data, which are synthetic images. The last 2 lines show the target (test) data, which are real images.}
    \label{fig:visda}
\end{figure}

\textbf{Datasets.} MNIST is a dataset of handwritten digits which contains 70,000 $28\times 28$ grayscale images.
MNIST-M is a modified version of MNIST, which contains colorful backgrounds extracted from color photos.
USPS dataset contains 9,298 $16\times 16$ pixel grayscale images.
SVHN dataset contains 600,000 $32\times 32$ RGB images of digits cropped from pictures of house number plates.
VisDA-C is a simulation-to-real dataset. The source domain contains the synthetic images rendered using 3D CAD models while the target domain contains the real images.
Some samples of the VisDA-C dataset are shown in Fig.~\ref{fig:visda}.

\textbf{Model and Baselines.}
In our research, we employ two types of models to analyze digits datasets: a simple convolutional neural network (CNN) and ResNet18.
To evaluate the effectiveness of ResNet18, we have conducted experiments utilizing both randomly initialized weights and ImageNet pretrained weights.
The architecture of the simple CNN is outlined in Algorithm~\ref{code:cnn}.
To ensure consistency in image size, all digit images have been resized to $64 \times 64$ during the training process. Additionally, for the MNIST and USPS datasets, we have duplicated the tensors to convert them into RGB images.
We also compare our models to established baseline models, including FADA~\cite{motiian2017few}, CCSA~\cite{motiian2017unified} and d-SNE~\cite{xu2019d}.
For the VisDA-C dataset, we utilize the ImageNet pretrained ResNet50.
All the images are resized to $224 \times 224$ for both training and testing.

\begin{algorithm}[ht]
\caption{PyTorch code of the simple CNN structure.}
\label{code:cnn}
\definecolor{codeblue}{rgb}{0.25,0.5,0.5}
\definecolor{codekw}{rgb}{0.85, 0.18, 0.50}
\lstset{
  backgroundcolor=\color{white},
  basicstyle=\fontsize{7.5pt}{7.5pt}\ttfamily\selectfont,
  columns=fullflexible,
  breaklines=true,
  captionpos=b,
  commentstyle=\fontsize{7.5pt}{7.5pt}\color{codeblue},
  keywordstyle=\fontsize{7.5pt}{7.5pt}\color{codekw},
}
\begin{lstlisting}[language=python]
# channels: number of channels for the input image
# num_classes: number of classes for the dataset. MNIST has 10 classes

class CNN(nn.Module):
    def __init__(self, channels, num_classes):
        super(CNN, self).__init__()
        self.conv = nn.Sequential(
            nn.Conv2d(channels, 32, kernel_size=5, stride=1, padding=2, bias=True),
            nn.ReLU(True),
            nn.MaxPool2d(kernel_size=2, stride=2, padding=0),
            nn.Conv2d(32, 48, kernel_size=5, stride=1, padding=2),
            nn.ReLU(True),
            nn.MaxPool2d(kernel_size=2, stride=2, padding=0),
        )
        self.fc = nn.Sequential(
            nn.Linear(12288, 100),
            nn.ReLU(True),
            nn.Linear(100, 100),
            nn.ReLU(True),
            nn.Linear(100, num_classes)
        )
\end{lstlisting}
\end{algorithm}

\textbf{Our Procedure.}
The initial steps for this procedure are comparable to those for out-of-distribution generalization, including initialization, multiple runs, and averaging of the model weights. 
For the initialization of the models on digits datasets, we make sure the shared initialization models achieve $\ge$ 85\% accuracy on their source domain.
However, there are two distinct strategies for the fourth adaptation step. 
The first strategy is to average the weights prior to adaptation and then test the final averaged model. 
The second strategy is to adapt each model individually, average their adapted weights, and then test. 
The results for both strategies are reported in our experiments, all of which involve launching 10 individual runs.
In the ablation studies, we also compare the effect of different optimizers.

\subsubsection{Digits Datasets Adaptation}
In this part we present the domain adaptation results using weight averaing on the digits datasets.
In the following content, we use $\mathcal{M}$, $\mathcal{MM}$, $\mathcal{U}$, $\mathcal{S}$ to denote \texttt{MNIST}, \texttt{MNIST-M}, \texttt{USPS} and \texttt{SVHN} respectively.

\textbf{Adaptation after weight averaging.}
The first strategy for this procedure is adaptation after weight averaging.
Specifically, after obtaining individually trained models, we will average them to create an averaged model. 
We will then adapt this model to the target domain and evaluate its performance on the test split of the target domain. 
These results are presented in Table~\ref{tab:digits_adapt_after_wa}. 
As can be seen from the table, weight averaging can significantly improve the performance of few-shot domain adaptation on digit datasets in most cases, particularly for \texttt{MNIST$\rightarrow$SVHN}.

\begin{table}[ht]
    \centering
    \begin{tabular}{p{1.8cm}|c|c|c|c|c|c}
    \midrule
    \textbf{Method} & \textbf{k} & $\mathcal{M}\rightarrow\mathcal{MM}$ & $\mathcal{M}\rightarrow\mathcal{U}$ & $\mathcal{M}\rightarrow\mathcal{S}$ & $\mathcal{U}\rightarrow\mathcal{M}$ &  $\mathcal{S}\rightarrow\mathcal{M}$ \\
    \midrule
    FADA~\cite{motiian2017few}  & 7  & - & 94.40 & 47.00 & 91.50 & 87.20\\
    CCSA~\cite{motiian2017unified} & 10 & 78.29 $\pm$ 2.00 & 92.27 $\pm$ 0.19 & 37.63 $\pm$ 3.62 & 95.71 $\pm$ 0.42 & 94.57 $\pm$ 0.40\\
    d-SNE~\cite{xu2019d} & 7  & $84.62 \pm 0.04$ & \underline{97.53} $\pm$ 0.10 & $53.19 \pm 0.28$ & \underline{97.52} $\pm$ 0.08 & 95.68 $\pm$ 0.03\\
    d-SNE~\cite{xu2019d} & 10 & $87.80 \pm 0.16$ & \textbf{99.00} $\pm$ 0.08 & $61.73 \pm 0.47$ & \textbf{98.49} $\pm$ 0.35 & \textbf{96.45} $\pm$ 0.20 \\
    \midrule
    \multirow{2}{1.8cm}{Adapt after WA} & 7   & \underline{84.78} & 95.42 & \underline{71.34} & 96.71 & \underline{95.91} \\
                         & 10 & \textbf{89.06} & 96.16 & \textbf{70.66} & 96.03 & 95.02\\
    \midrule
    Adapt before WA & 10   & 69.33 & 92.78 & 59.40 & 95.24 & 89.54 \\
    \midrule
    \end{tabular}
    \caption{Digits datasets adaptation after weight averaging. The best results for 10-shot adaptation are highlighted in \textbf{bold} and the best results for 7-shot adaptation are highlighted with \underline{underlines}.}
    \label{tab:digits_adapt_after_wa}
\end{table}


\textbf{Adaptation before weight averaging.}
The second strategy we will employ is adaptation before weight averaging. Instead of averaging all the weights to obtain the final model, we will adapt each individual model to the target domain separately. Once all the models have been adapted, we will then average them and test the resulting averaged model directly on the test split of the target domain. The results of this strategy can be seen in Table~\ref{tab:digits_adapt_after_wa}. 
However, it is clear that this approach results in worse performance than adaptation after weight averaging. We believe this is due to the destruction of learned common features during individual adaptation. As a result, we will not use this approach in the rest of our analysis and will only report results for adaptation after weight averaging.

\subsubsection{Ablation Study: The Effect of Model Architecture}
In this part, we investigate the impact of model architecture on the domain adaptation performance of weight averaging through ablation studies.
We compare 2 models: simple CNN and ResNet18.
For ResNet18, we also examine the effect of random initialization and ImageNet pretrained initialization.
The comparison results are presented in Table~\ref{tab:ablation_models}.
It should be noted that the results for the simple 2-layer CNN on \texttt{SVHN$\rightarrow$MNIST} adaptation task are not reported as the model was not able to fit the \texttt{SVHN} dataset. 
The results in the table show that the ImageNet-pretrained ResNet18 consistently achieves the best performance across all adaptation tasks.

\begin{table}[ht]
    \centering
    \begin{tabular}{c|c|c|c|c|c|c}
        \hline
        \textbf{Model} & \textbf{Init.} & $\mathcal{M}\rightarrow\mathcal{MM}$ & $\mathcal{M}\rightarrow\mathcal{U}$ & $\mathcal{M}\rightarrow\mathcal{S}$ & $\mathcal{U}\rightarrow\mathcal{M}$ &  $\mathcal{S}\rightarrow\mathcal{M}$ \\
        \midrule
        CNN      & Random   & 73.03 & 90.63 & 48.29 & 91.59 & - \\
        ResNet18 & Random   & 73.73 & 94.32 & 58.45 & 93.19 & 94.07 \\
        ResNet18 & ImageNet & \textbf{89.06} & \textbf{96.16} & \textbf{70.66} & \textbf{96.03} & \textbf{95.02} \\
        \midrule
    \end{tabular}
    \caption{Results of ablation study on the effect of model architectures on the domain adaptation performance of weight averaging. All the results are 10-shot adaptation. All the models are optimized with SAM optimizer.}
    \label{tab:ablation_models}
\end{table}

\subsubsection{Ablation Study: The Effect of Optimizer}
In this section, we evaluate the impact of different optimizers on domain adaptation performance in the context of weight averaging. Specifically, we compare Adam and Sharpness-Aware Minimization (SAM). We report results for 10-shot adaptation and the results are listed in Table~\ref{tab:ablation_optimizer}. Interestingly, we observe that the SAM optimizer consistently leads to better few-shot domain adaptation performance across all models and adaptation tasks. This highlights the superiority of the SAM optimizer for this task.

\begin{table}[ht]
    \centering
    \begin{tabular}{c|c|c|c|c|c|c}
        \hline
        \textbf{Model} & \textbf{Optimizer} & $\mathcal{M}\rightarrow\mathcal{MM}$ & $\mathcal{M}\rightarrow\mathcal{U}$ & $\mathcal{M}\rightarrow\mathcal{S}$ & $\mathcal{U}\rightarrow\mathcal{M}$ &  $\mathcal{S}\rightarrow\mathcal{M}$ \\
        \midrule
        CNN & Adam & 69.07 & 91.93 & 47.50 & 89.88 & - \\
        CNN & SAM  & 73.03 & 90.63 & 48.29 & 91.59 & - \\
        \midrule
        ResNet18 & Adam & 64.58 & 91.63 & 51.67 & 93.07 & 94.04 \\
        ResNet18 & SAM  & 73.73 & 94.32 & 58.45 & 93.19 & 94.07 \\
        \midrule
        ResNet18 (ImageNet) & Adam & 83.56 & 94.82 & 65.15 & 95.27 & 94.97 \\
        ResNet18 (ImageNet) & SAM  & 89.06 & 96.16 & 70.66 & 96.03 & 95.02 \\
        \midrule
    \end{tabular}
    \caption{Results of ablation study on the effect of optimizers on the domain adaptation performance of weight averaging. All the results are 10-shot adaptation.}
    \label{tab:ablation_optimizer}
\end{table}

\subsubsection{Ablation Study: The Effect of The Number of Adapted Samples}
\begin{figure}[ht]
    \centering
    \includegraphics[width=\textwidth]{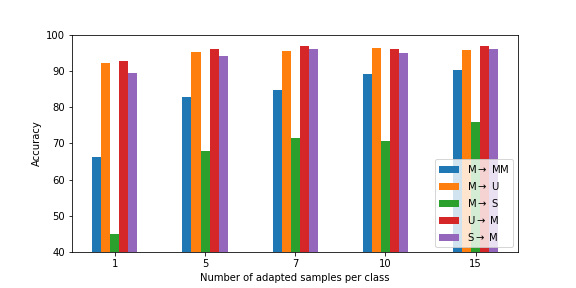}
    \caption{The effect of the number of adapted samples per class.}
    \label{fig:num_adapt}
\end{figure}

In this part, we investigate the effect of the number of samples used for adaptation on performance. We have previously shown that the SAM optimizer and an ImageNet-pretrained ResNet18 model provide superior results. Therefore, in this analysis, we only report results using these configurations. The results are presented in Fig.~\ref{fig:num_adapt}. These results demonstrate how the performance of the model changes as the number of samples used for adaptation increases.
As can be seen from the figure, weight averaging for few-shot domain adaptation is quite data-efficient.
The accuracy values are already high when only 5 samples per class are used.
Also we notice that the accuracy of the model increases consistently as the number of samples used for adaptation increases. However, the rate of increase becomes less drastic as the number of samples increases. The trend can be observed as a gentle increase in accuracy.

\subsubsection{Ablation Study: t-SNE Feature Analysis}

\begin{figure}[ht]
     \centering
     \begin{subfigure}[b]{0.32\textwidth}
         \centering
         \includegraphics[width=\textwidth]{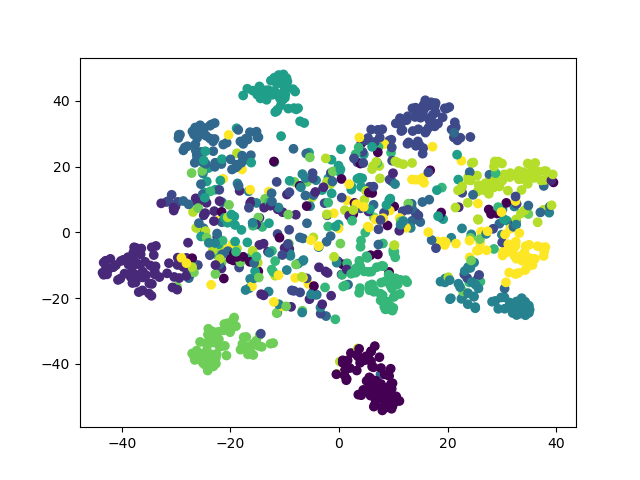}
     \end{subfigure}
     \hfill
     \begin{subfigure}[b]{0.32\textwidth}
         \centering
         \includegraphics[width=\textwidth]{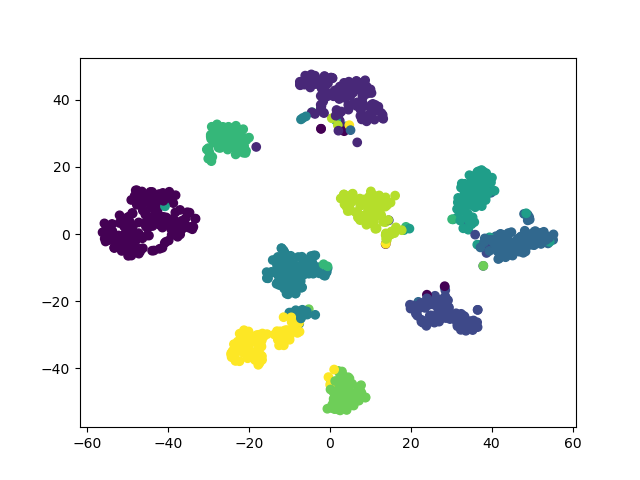}
     \end{subfigure}
     \hfill
     \begin{subfigure}[b]{0.32\textwidth}
         \centering
         \includegraphics[width=\textwidth]{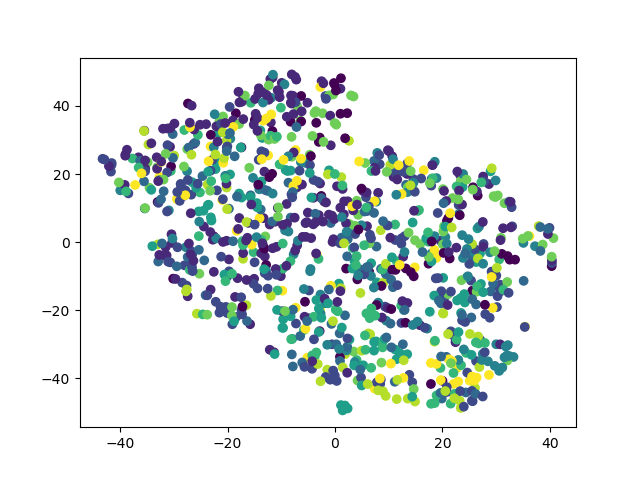}
     \end{subfigure}

    \begin{subfigure}[b]{0.32\textwidth}
         \centering
         \includegraphics[width=\textwidth]{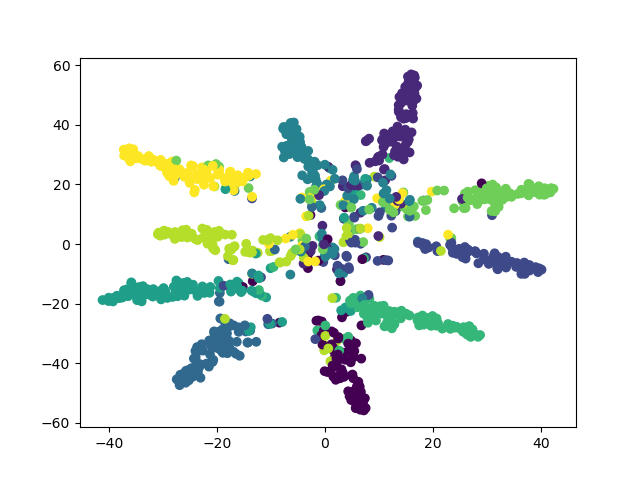}
     \end{subfigure}
     \hfill
     \begin{subfigure}[b]{0.32\textwidth}
         \centering
         \includegraphics[width=\textwidth]{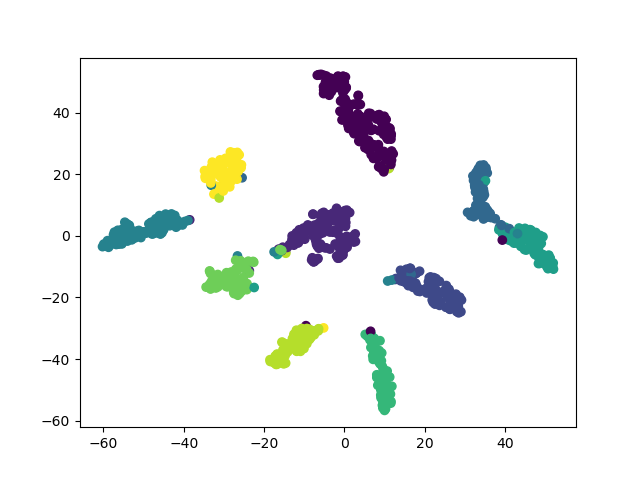}
     \end{subfigure}
     \hfill
     \begin{subfigure}[b]{0.32\textwidth}
         \centering
         \includegraphics[width=\textwidth]{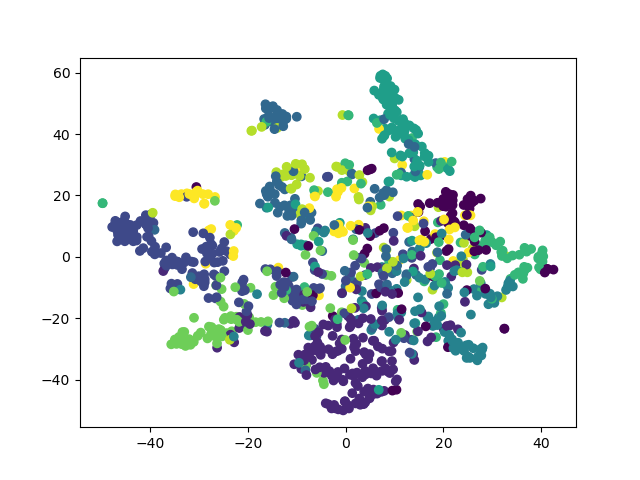}
     \end{subfigure}

     \begin{subfigure}[b]{0.32\textwidth}
         \centering
         \includegraphics[width=\textwidth]{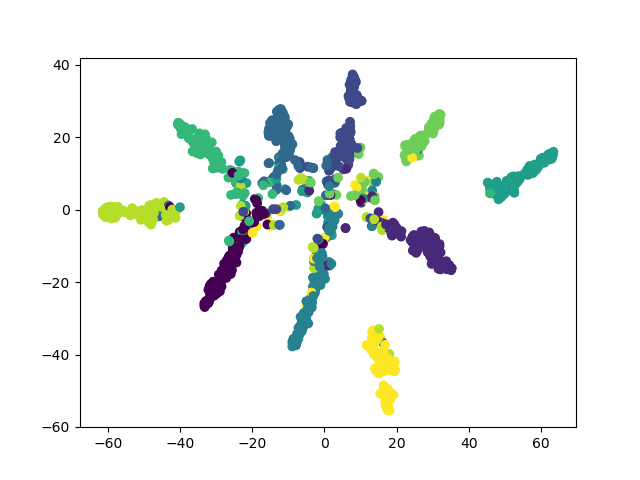}
         \caption{MNIST $\rightarrow$ MNIST-M}
     \end{subfigure}
     \hfill
     \begin{subfigure}[b]{0.32\textwidth}
         \centering
         \includegraphics[width=\textwidth]{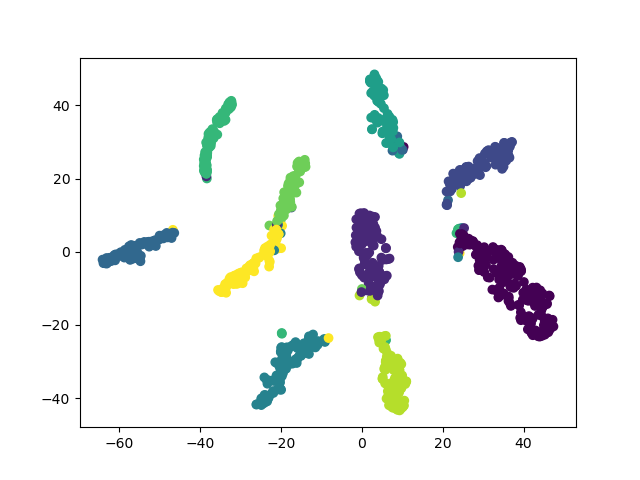}
         \caption{MNIST $\rightarrow$ USPS}
     \end{subfigure}
     \hfill
     \begin{subfigure}[b]{0.32\textwidth}
         \centering
         \includegraphics[width=\textwidth]{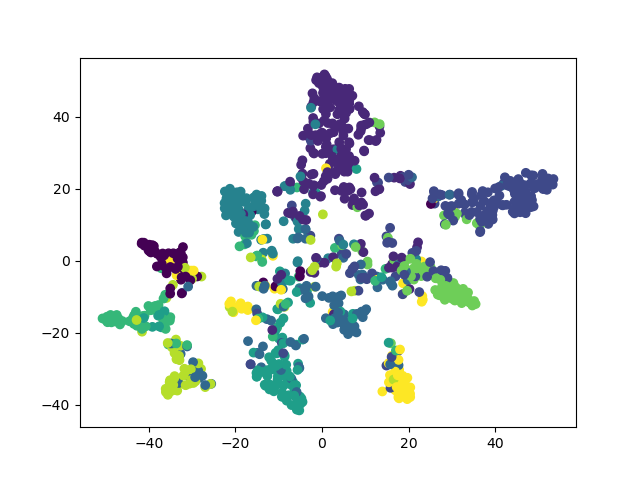}
         \caption{MNIST $\rightarrow$ SVHN}
     \end{subfigure}
        
        \caption{t-SNE visualization for ResNet18 (ImageNet pretrained). The top row shows the features of a well-trained ResNet18 on MNIST without few-shot adaptation.
        The middle row shows the features of the above well-trained ResNet18 after 10-shot domain adatation.
        The bottom row shows the features of the weight averaged ResNet18 after 10-shot domain adaptation. All the models are ImageNet pretrained and optimized with SAM.}
        \label{fig:tsne_res18_imgnet_sam}
\end{figure}

To gain a deeper understanding of the effect of weight averaging on domain adaptation performance, we visualize the features of classifiers before and after adaptation. 
Specifically, we compare a well-trained model with and without adaptation to a weight-averaged model after adaptation. 
The features are extracted from the final \texttt{fc} layer of each classifier. We use t-SNE~\cite{van2008visualizing} to visualize the features. The results are shown in Fig.~\ref{fig:tsne_res18_imgnet_sam}. By comparing the three rows in the figure, we can clearly observe that weight averaging greatly improves the performance of few-shot domain adaptation.

\newpage
\subsubsection{VisDA-C Adaptation}
In this section, we present the results of utilizing weight averaging for domain adaptation on the VisDA-C dataset. 
Our approach is compared against d-SNE~\cite{xu2019d} and LCCS (Linear Combination Coefficients for Batch Normalization Statistics)~\cite{zhang2022few}. 
The experiments were conducted using the SAM optimizer with a total of 10 runs, and the adaptation strategy implemented is adaptation following weight averaging. 
The results of the experiments are presented in Table~\ref{tab:visda_result}.

\begin{table}[ht]
    \footnotesize
    \centering
    \begin{tabular}{c|c|c}
        \midrule
        \textbf{Method} & \textbf{k} & \textbf{VisDA Real} \\
        \midrule
        LCCS~\cite{zhang2022few} & 10 & 79.2 \\
        d-SNE~\cite{xu2019d} & 10 & 80.66 \\
        \midrule
        WA & 10 & 49.41 \\
        \midrule
    \end{tabular}
    \caption{VisDA Weight Averaging before Few-Shot Adaptation Results}
    \label{tab:visda_result}
\end{table}

Our current methodology exhibits a significant discrepancy when compared to state-of-the-art techniques. In order to thoroughly evaluate our results, additional analysis and experimentation will be conducted.

\clearpage
\section{Discussion}
\label{sec:thesis_discussion}

Distribution shift is a critical issue in machine learning and computer vision.
In this thesis, we tacked two related problems in this area, specifically out-of-distribution generalization and few-shot domain adaptation.
To address the issue of OOD generalization, we proposed an extension to existing weight averaging-based methods by incorporating gradient similarity as a regularizer to enforce model diversity.
In the context of few-shot domain adaptation, we demonstrated the effectiveness of weight averaging through extensive empirical evaluations.
Our findings suggest that weight averaging is a viable approach for addressing distribution shift in machine learning and computer vision.

However, there are also some limitations and potential problems in this thesis.
To provide a comprehensive understanding of our work, we include several discussions to help the reader interpret our results and motivate future research.

\textbf{What is best way to increase model diversity in weight averaging?} In this thesis, we explored the use of weight averaging as a method for improving generalization performance in machine learning. 
Previous research has shown that averaging the weights of diversified models can lead to better results. 
However, simply relying on using different hyperparameters during training is not sufficient for achieving optimal diversity.
To address this issue, we experimented with incorporating gradient diversity as a means of enforcing model diversity. While the results of our experiments were not entirely promising, we believe that explicitly enforcing model diversity during training is a crucial factor for achieving better generalization. However, the optimal method for achieving this is still an open question and requires further investigation.

\textbf{Is weight averaging uniformly useful for all downstream tasks?}
In this thesis, we focused on the application of weight averaging in the context of computer vision tasks, specifically image classification. However, it is important to note that this method may also be applicable to other downstream tasks such as image segmentation and object detection. Further exploration is needed to evaluate the effectiveness of weight averaging for these tasks in a more comprehensive manner. This could be a potential avenue for future research.

\textbf{Weight averaging is computationally intensive.} 
Weight averaging is a method that relies on training multiple individual models, which can be computationally intensive. For simple tasks and small models, this requirement is manageable and affordable. However, in practical scenarios where the problem is complex and there is a large amount of data and very large models, weight averaging may not be efficient and may not be a long-term solution. Despite the promising results that weight averaging has shown in out-of-distribution generalization, it is important to consider the computational cost and limitations when applying this method in practice.\clearpage

\cleardoublepage
\bibliographystyle{plain}
\bibliography{thesis}

\begin{thebibliography}{100}

\bibitem{albuquerque2020adversarial}
Isabela Albuquerque, João Monteiro, Tiago~H. Falk, and Ioannis Mitliagkas.
\newblock Adversarial target-invariant representation learning for domain generalization.
\newblock {\em CoRR}, abs/1911.00804, 2019.

\bibitem{albuquerque2020improving}
Isabela Albuquerque, Nikhil Naik, Junnan Li, Nitish Keskar, and Richard Socher.
\newblock Improving out-of-distribution generalization via multi-task self-supervised pretraining.
\newblock {\em arXiv preprint arXiv:2003.13525}, 2020.

\bibitem{andriushchenko2022towards}
Maksym Andriushchenko and Nicolas Flammarion.
\newblock Towards understanding sharpness-aware minimization.
\newblock In {\em International Conference on Machine Learning}, pages 639--668. PMLR, 2022.

\bibitem{arpit2021ensemble}
Devansh Arpit, Huan Wang, Yingbo Zhou, and Caiming Xiong.
\newblock Ensemble of averages: Improving model selection and boosting performance in domain generalization.
\newblock {\em arXiv preprint arXiv:2110.10832}, 2021.

\bibitem{barbu2019objectnet}
Andrei Barbu, David Mayo, Julian Alverio, William Luo, Christopher Wang, Dan Gutfreund, Josh Tenenbaum, and Boris Katz.
\newblock Objectnet: A large-scale bias-controlled dataset for pushing the limits of object recognition models.
\newblock {\em Advances in neural information processing systems}, 32, 2019.

\bibitem{bauer1999empirical}
Eric Bauer and Ron Kohavi.
\newblock An empirical comparison of voting classification algorithms: Bagging, boosting, and variants.
\newblock {\em Machine learning}, 36(1):105--139, 1999.

\bibitem{beede2020human}
Emma Beede, Elizabeth Baylor, Fred Hersch, Anna Iurchenko, Lauren Wilcox, Paisan Ruamviboonsuk, and Laura~M Vardoulakis.
\newblock A human-centered evaluation of a deep learning system deployed in clinics for the detection of diabetic retinopathy.
\newblock In {\em Proceedings of the 2020 CHI conference on human factors in computing systems}, pages 1--12, 2020.

\bibitem{beery2018recognition}
Sara Beery, Grant Van~Horn, and Pietro Perona.
\newblock Recognition in terra incognita.
\newblock In {\em Proceedings of the European conference on computer vision (ECCV)}, pages 456--473, 2018.

\bibitem{bommasani2021opportunities}
Rishi Bommasani, Drew~A Hudson, Ehsan Adeli, Russ Altman, Simran Arora, Sydney von Arx, Michael~S Bernstein, Jeannette Bohg, Antoine Bosselut, Emma Brunskill, et~al.
\newblock On the opportunities and risks of foundation models.
\newblock {\em arXiv preprint arXiv:2108.07258}, 2021.

\bibitem{breiman1996bagging}
Leo Breiman.
\newblock Bagging predictors.
\newblock {\em Machine learning}, 24(2):123--140, 1996.

\bibitem{breiman2001random}
Leo Breiman.
\newblock Random forests.
\newblock {\em Machine learning}, 45(1):5--32, 2001.

\bibitem{brown2020language}
Tom Brown, Benjamin Mann, Nick Ryder, Melanie Subbiah, Jared~D Kaplan, Prafulla Dhariwal, Arvind Neelakantan, Pranav Shyam, Girish Sastry, Amanda Askell, et~al.
\newblock Language models are few-shot learners.
\newblock {\em Advances in neural information processing systems}, 33:1877--1901, 2020.

\bibitem{cha2021swad}
Junbum Cha, Sanghyuk Chun, Kyungjae Lee, Han-Cheol Cho, Seunghyun Park, Yunsung Lee, and Sungrae Park.
\newblock Swad: Domain generalization by seeking flat minima.
\newblock {\em Advances in Neural Information Processing Systems}, 34:22405--22418, 2021.

\bibitem{choi2010exploiting}
Myung~Jin Choi, Joseph~J Lim, Antonio Torralba, and Alan~S Willsky.
\newblock Exploiting hierarchical context on a large database of object categories.
\newblock In {\em 2010 IEEE computer society conference on computer vision and pattern recognition}, pages 129--136. IEEE, 2010.

\bibitem{deng2009imagenet}
Jia Deng, Wei Dong, Richard Socher, Li-Jia Li, Kai Li, and Li~Fei-Fei.
\newblock Imagenet: A large-scale hierarchical image database.
\newblock In {\em 2009 IEEE conference on computer vision and pattern recognition}, pages 248--255. Ieee, 2009.

\bibitem{devlin2018bert}
Jacob Devlin, Ming-Wei Chang, Kenton Lee, and Kristina Toutanova.
\newblock Bert: Pre-training of deep bidirectional transformers for language understanding.
\newblock {\em arXiv preprint arXiv:1810.04805}, 2018.

\bibitem{dinh2017sharp}
Laurent Dinh, Razvan Pascanu, Samy Bengio, and Yoshua Bengio.
\newblock Sharp minima can generalize for deep nets.
\newblock In {\em International Conference on Machine Learning}, pages 1019--1028. PMLR, 2017.

\bibitem{dittadi2021role}
Andrea Dittadi, Frederik Tr{\"a}uble, Manuel W{\"u}thrich, Felix Widmaier, Peter Gehler, Ole Winther, Francesco Locatello, Olivier Bachem, Bernhard Sch{\"o}lkopf, and Stefan Bauer.
\newblock The role of pretrained representations for the ood generalization of rl agents.
\newblock {\em arXiv preprint arXiv:2107.05686}, 2021.

\bibitem{everingham2010pascal}
Mark Everingham, Luc Van~Gool, Christopher~KI Williams, John Winn, and Andrew Zisserman.
\newblock The pascal visual object classes (voc) challenge.
\newblock {\em International journal of computer vision}, 88(2):303--338, 2010.

\bibitem{fang2013unbiased}
Chen Fang, Ye~Xu, and Daniel~N Rockmore.
\newblock Unbiased metric learning: On the utilization of multiple datasets and web images for softening bias.
\newblock In {\em Proceedings of the IEEE International Conference on Computer Vision}, pages 1657--1664, 2013.

\bibitem{fei2004learning}
Li~Fei-Fei, Rob Fergus, and Pietro Perona.
\newblock Learning generative visual models from few training examples: An incremental bayesian approach tested on 101 object categories.
\newblock In {\em 2004 conference on computer vision and pattern recognition workshop}, pages 178--178. IEEE, 2004.

\bibitem{foret2020sharpness}
Pierre Foret, Ariel Kleiner, Hossein Mobahi, and Behnam Neyshabur.
\newblock Sharpness-aware minimization for efficiently improving generalization.
\newblock {\em arXiv preprint arXiv:2010.01412}, 2020.

\bibitem{frankle2020linear}
Jonathan Frankle, Gintare~Karolina Dziugaite, Daniel Roy, and Michael Carbin.
\newblock Linear mode connectivity and the lottery ticket hypothesis.
\newblock In {\em International Conference on Machine Learning}, pages 3259--3269. PMLR, 2020.

\bibitem{freund1997decision}
Yoav Freund and Robert~E Schapire.
\newblock A decision-theoretic generalization of on-line learning and an application to boosting.
\newblock {\em Journal of computer and system sciences}, 55(1):119--139, 1997.

\bibitem{ganin2015unsupervised}
Yaroslav Ganin and Victor Lempitsky.
\newblock Unsupervised domain adaptation by backpropagation.
\newblock In {\em International conference on machine learning}, pages 1180--1189. PMLR, 2015.

\bibitem{ganin2016domain}
Yaroslav Ganin, Evgeniya Ustinova, Hana Ajakan, Pascal Germain, Hugo Larochelle, Fran{\c{c}}ois Laviolette, Mario Marchand, and Victor Lempitsky.
\newblock Domain-adversarial training of neural networks.
\newblock {\em The journal of machine learning research}, 17(1):2096--2030, 2016.

\bibitem{geirhos2020shortcut}
Robert Geirhos, J{\"o}rn-Henrik Jacobsen, Claudio Michaelis, Richard Zemel, Wieland Brendel, Matthias Bethge, and Felix~A Wichmann.
\newblock Shortcut learning in deep neural networks.
\newblock {\em Nature Machine Intelligence}, 2(11):665--673, 2020.

\bibitem{gulrajani2020search}
Ishaan Gulrajani and David Lopez-Paz.
\newblock In search of lost domain generalization.
\newblock {\em arXiv preprint arXiv:2007.01434}, 2020.

\bibitem{guo2022stochastic}
Hao Guo, Jiyong Jin, and Bin Liu.
\newblock Stochastic weight averaging revisited.
\newblock {\em arXiv preprint arXiv:2201.00519}, 2022.

\bibitem{gupta2020stochastic}
Vipul Gupta, Santiago~Akle Serrano, and Dennis DeCoste.
\newblock Stochastic weight averaging in parallel: Large-batch training that generalizes well.
\newblock {\em arXiv preprint arXiv:2001.02312}, 2020.

\bibitem{he2020momentum}
Kaiming He, Haoqi Fan, Yuxin Wu, Saining Xie, and Ross Girshick.
\newblock Momentum contrast for unsupervised visual representation learning.
\newblock In {\em Proceedings of the IEEE/CVF conference on computer vision and pattern recognition}, pages 9729--9738, 2020.

\bibitem{he2016deep}
Kaiming He, Xiangyu Zhang, Shaoqing Ren, and Jian Sun.
\newblock Deep residual learning for image recognition.
\newblock In {\em Proceedings of the IEEE conference on computer vision and pattern recognition}, pages 770--778, 2016.

\bibitem{hendrycks2019benchmarking}
Dan Hendrycks and Thomas Dietterich.
\newblock Benchmarking neural network robustness to common corruptions and perturbations.
\newblock {\em arXiv preprint arXiv:1903.12261}, 2019.

\bibitem{hermann2020shapes}
Katherine Hermann and Andrew Lampinen.
\newblock What shapes feature representations? exploring datasets, architectures, and training.
\newblock {\em Advances in Neural Information Processing Systems}, 33:9995--10006, 2020.

\bibitem{higgins2017betavae}
Irina Higgins, Loic Matthey, Arka Pal, Christopher Burgess, Xavier Glorot, Matthew Botvinick, Shakir Mohamed, and Alexander Lerchner.
\newblock beta-{VAE}: Learning basic visual concepts with a constrained variational framework.
\newblock In {\em International Conference on Learning Representations}, 2017.

\bibitem{ho2020denoising}
Jonathan Ho, Ajay Jain, and Pieter Abbeel.
\newblock Denoising diffusion probabilistic models.
\newblock {\em Advances in Neural Information Processing Systems}, 33:6840--6851, 2020.

\bibitem{hochreiter1994simplifying}
Sepp Hochreiter and J{\"u}rgen Schmidhuber.
\newblock Simplifying neural nets by discovering flat minima.
\newblock {\em Advances in neural information processing systems}, 7, 1994.

\bibitem{hull1994database}
Jonathan~J. Hull.
\newblock A database for handwritten text recognition research.
\newblock {\em IEEE Transactions on pattern analysis and machine intelligence}, 16(5):550--554, 1994.

\bibitem{ilyas2019adversarial}
Andrew Ilyas, Shibani Santurkar, Dimitris Tsipras, Logan Engstrom, Brandon Tran, and Aleksander Madry.
\newblock Adversarial examples are not bugs, they are features.
\newblock {\em Advances in neural information processing systems}, 32, 2019.

\bibitem{izmailov2018averaging}
Pavel Izmailov, Dmitrii Podoprikhin, Timur Garipov, Dmitry Vetrov, and Andrew~Gordon Wilson.
\newblock Averaging weights leads to wider optima and better generalization.
\newblock {\em arXiv preprint arXiv:1803.05407}, 2018.

\bibitem{jia2021scaling}
Chao Jia, Yinfei Yang, Ye~Xia, Yi-Ting Chen, Zarana Parekh, Hieu Pham, Quoc Le, Yun-Hsuan Sung, Zhen Li, and Tom Duerig.
\newblock Scaling up visual and vision-language representation learning with noisy text supervision.
\newblock In {\em International Conference on Machine Learning}, pages 4904--4916. PMLR, 2021.

\bibitem{kaddour2022questions}
Jean Kaddour, Linqing Liu, Ricardo Silva, and Matt~J Kusner.
\newblock When do flat minima optimizers work?
\newblock {\em arXiv preprint arXiv:2202.00661}, 2022.

\bibitem{keskar2016large}
Nitish~Shirish Keskar, Dheevatsa Mudigere, Jorge Nocedal, Mikhail Smelyanskiy, and Ping Tak~Peter Tang.
\newblock On large-batch training for deep learning: Generalization gap and sharp minima.
\newblock {\em arXiv preprint arXiv:1609.04836}, 2016.

\bibitem{kilbertus2018generalization}
Niki Kilbertus, Giambattista Parascandolo, and Bernhard Sch{\"o}lkopf.
\newblock Generalization in anti-causal learning.
\newblock {\em arXiv preprint arXiv:1812.00524}, 2018.

\bibitem{kim2018disentangling}
Hyunjik Kim and Andriy Mnih.
\newblock Disentangling by factorising.
\newblock In {\em International Conference on Machine Learning}, pages 2649--2658. PMLR, 2018.

\bibitem{kirichenko2022last}
Polina Kirichenko, Pavel Izmailov, and Andrew~Gordon Wilson.
\newblock Last layer re-training is sufficient for robustness to spurious correlations.
\newblock {\em arXiv preprint arXiv:2204.02937}, 2022.

\bibitem{kolesnikov2020big}
Alexander Kolesnikov, Lucas Beyer, Xiaohua Zhai, Joan Puigcerver, Jessica Yung, Sylvain Gelly, and Neil Houlsby.
\newblock Big transfer (bit): General visual representation learning.
\newblock In {\em European conference on computer vision}, pages 491--507. Springer, 2020.

\bibitem{krizhevsky2009learning}
Alex Krizhevsky, Geoffrey Hinton, et~al.
\newblock Learning multiple layers of features from tiny images, 2009.

\bibitem{kumar2022fine}
Ananya Kumar, Aditi Raghunathan, Robbie Jones, Tengyu Ma, and Percy Liang.
\newblock Fine-tuning can distort pretrained features and underperform out-of-distribution.
\newblock {\em arXiv preprint arXiv:2202.10054}, 2022.

\bibitem{kwon2021asam}
Jungmin Kwon, Jeongseop Kim, Hyunseo Park, and In~Kwon Choi.
\newblock Asam: Adaptive sharpness-aware minimization for scale-invariant learning of deep neural networks.
\newblock In {\em International Conference on Machine Learning}, pages 5905--5914. PMLR, 2021.

\bibitem{lecun1998gradient}
Yann LeCun, L{\'e}on Bottou, Yoshua Bengio, and Patrick Haffner.
\newblock Gradient-based learning applied to document recognition.
\newblock {\em Proceedings of the IEEE}, 86(11):2278--2324, 1998.

\bibitem{li2017deeper}
Da~Li, Yongxin Yang, Yi-Zhe Song, and Timothy~M Hospedales.
\newblock Deeper, broader and artier domain generalization.
\newblock In {\em Proceedings of the IEEE international conference on computer vision}, pages 5542--5550, 2017.

\bibitem{li2021cross}
Jichang Li, Guanbin Li, Yemin Shi, and Yizhou Yu.
\newblock Cross-domain adaptive clustering for semi-supervised domain adaptation.
\newblock In {\em Proceedings of the IEEE/CVF Conference on Computer Vision and Pattern Recognition}, pages 2505--2514, 2021.

\bibitem{li2021few}
Shaohua Li, Xiuchao Sui, Jie Fu, Huazhu Fu, Xiangde Luo, Yangqin Feng, Xinxing Xu, Yong Liu, Daniel~SW Ting, and Rick Siow~Mong Goh.
\newblock Few-shot domain adaptation with polymorphic transformers.
\newblock In {\em International Conference on Medical Image Computing and Computer-Assisted Intervention}, pages 330--340. Springer, 2021.

\bibitem{li2018domain}
Shuang Li, Shiji Song, Gao Huang, Zhengming Ding, and Cheng Wu.
\newblock Domain invariant and class discriminative feature learning for visual domain adaptation.
\newblock {\em IEEE transactions on image processing}, 27(9):4260--4273, 2018.

\bibitem{long2015learning}
Mingsheng Long, Yue Cao, Jianmin Wang, and Michael Jordan.
\newblock Learning transferable features with deep adaptation networks.
\newblock In {\em International conference on machine learning}, pages 97--105. PMLR, 2015.

\bibitem{long2017deep}
Mingsheng Long, Han Zhu, Jianmin Wang, and Michael~I Jordan.
\newblock Deep transfer learning with joint adaptation networks.
\newblock In {\em International conference on machine learning}, pages 2208--2217. PMLR, 2017.

\bibitem{michaelis2019benchmarking}
Claudio Michaelis, Benjamin Mitzkus, Robert Geirhos, Evgenia Rusak, Oliver Bringmann, Alexander~S Ecker, Matthias Bethge, and Wieland Brendel.
\newblock Benchmarking robustness in object detection: Autonomous driving when winter is coming.
\newblock {\em arXiv preprint arXiv:1907.07484}, 2019.

\bibitem{motiian2017few}
Saeid Motiian, Quinn Jones, Seyed Iranmanesh, and Gianfranco Doretto.
\newblock Few-shot adversarial domain adaptation.
\newblock {\em Advances in neural information processing systems}, 30, 2017.

\bibitem{motiian2017unified}
Saeid Motiian, Marco Piccirilli, Donald~A Adjeroh, and Gianfranco Doretto.
\newblock Unified deep supervised domain adaptation and generalization.
\newblock In {\em Proceedings of the IEEE international conference on computer vision}, pages 5715--5725, 2017.

\bibitem{muandet2013domain}
Krikamol Muandet, David Balduzzi, and Bernhard Sch{\"o}lkopf.
\newblock Domain generalization via invariant feature representation.
\newblock In {\em International Conference on Machine Learning}, pages 10--18. PMLR, 2013.

\bibitem{mustafa2020deep}
Basil Mustafa, Carlos Riquelme, Joan Puigcerver, Andr{\'e}~Susano Pinto, Daniel Keysers, and Neil Houlsby.
\newblock Deep ensembles for low-data transfer learning.
\newblock {\em arXiv preprint arXiv:2010.06866}, 2020.

\bibitem{nagarajan2020understanding}
Vaishnavh Nagarajan, Anders Andreassen, and Behnam Neyshabur.
\newblock Understanding the failure modes of out-of-distribution generalization.
\newblock {\em arXiv preprint arXiv:2010.15775}, 2020.

\bibitem{netzer2011reading}
Yuval Netzer, Tao Wang, Adam Coates, Alessandro Bissacco, Baolin Wu, Andrew~Y Ng, et~al.
\newblock Reading digits in natural images with unsupervised feature learning.
\newblock In {\em NIPS workshop on deep learning and unsupervised feature learning}, volume 2011, page~4. Granada, 2011.

\bibitem{neyshabur2017exploring}
Behnam Neyshabur, Srinadh Bhojanapalli, David McAllester, and Nati Srebro.
\newblock Exploring generalization in deep learning.
\newblock {\em Advances in neural information processing systems}, 30, 2017.

\bibitem{neyshabur2020being}
Behnam Neyshabur, Hanie Sedghi, and Chiyuan Zhang.
\newblock What is being transferred in transfer learning?
\newblock {\em Advances in neural information processing systems}, 33:512--523, 2020.

\bibitem{neyshabur2014search}
Behnam Neyshabur, Ryota Tomioka, and Nathan Srebro.
\newblock In search of the real inductive bias: On the role of implicit regularization in deep learning.
\newblock {\em arXiv preprint arXiv:1412.6614}, 2014.

\bibitem{ovadia2019can}
Yaniv Ovadia, Emily Fertig, Jie Ren, Zachary Nado, David Sculley, Sebastian Nowozin, Joshua Dillon, Balaji Lakshminarayanan, and Jasper Snoek.
\newblock Can you trust your model's uncertainty? evaluating predictive uncertainty under dataset shift.
\newblock {\em Advances in neural information processing systems}, 32, 2019.

\bibitem{peng2019moment}
Xingchao Peng, Qinxun Bai, Xide Xia, Zijun Huang, Kate Saenko, and Bo~Wang.
\newblock Moment matching for multi-source domain adaptation.
\newblock In {\em Proceedings of the IEEE/CVF international conference on computer vision}, pages 1406--1415, 2019.

\bibitem{visda2017}
Xingchao Peng, Ben Usman, Neela Kaushik, Judy Hoffman, Dequan Wang, and Kate Saenko.
\newblock Visda: The visual domain adaptation challenge, 2017.

\bibitem{radford2021learning}
Alec Radford, Jong~Wook Kim, Chris Hallacy, Aditya Ramesh, Gabriel Goh, Sandhini Agarwal, Girish Sastry, Amanda Askell, Pamela Mishkin, Jack Clark, et~al.
\newblock Learning transferable visual models from natural language supervision.
\newblock In {\em International Conference on Machine Learning}, pages 8748--8763. PMLR, 2021.

\bibitem{rame2022diverse}
Alexandre Rame, Matthieu Kirchmeyer, Thibaud Rahier, Alain Rakotomamonjy, Patrick Gallinari, and Matthieu Cord.
\newblock Diverse weight averaging for out-of-distribution generalization.
\newblock {\em arXiv preprint arXiv:2205.09739}, 2022.

\bibitem{ross2020ensembles}
Andrew Ross, Weiwei Pan, Leo Celi, and Finale Doshi-Velez.
\newblock Ensembles of locally independent prediction models.
\newblock In {\em Proceedings of the AAAI Conference on Artificial Intelligence}, volume~34, pages 5527--5536, 2020.

\bibitem{ruan2021optimal}
Yangjun Ruan, Yann Dubois, and Chris~J Maddison.
\newblock Optimal representations for covariate shift.
\newblock {\em arXiv preprint arXiv:2201.00057}, 2021.

\bibitem{russell2008labelme}
Bryan~C Russell, Antonio Torralba, Kevin~P Murphy, and William~T Freeman.
\newblock Labelme: a database and web-based tool for image annotation.
\newblock {\em International journal of computer vision}, 77(1):157--173, 2008.

\bibitem{ryu-etal-2018-domain}
Seonghan Ryu, Sangjun Koo, Hwanjo Yu, and Gary~Geunbae Lee.
\newblock Out-of-domain detection based on generative adversarial network.
\newblock In {\em Proceedings of the 2018 Conference on Empirical Methods in Natural Language Processing}, pages 714--718, Brussels, Belgium, October-November 2018. Association for Computational Linguistics.

\bibitem{sagi2018ensemble}
Omer Sagi and Lior Rokach.
\newblock Ensemble learning: A survey.
\newblock {\em Wiley Interdisciplinary Reviews: Data Mining and Knowledge Discovery}, 8(4):e1249, 2018.

\bibitem{saito2019semi}
Kuniaki Saito, Donghyun Kim, Stan Sclaroff, Trevor Darrell, and Kate Saenko.
\newblock Semi-supervised domain adaptation via minimax entropy.
\newblock In {\em Proceedings of the IEEE/CVF International Conference on Computer Vision}, pages 8050--8058, 2019.

\bibitem{saito2018maximum}
Kuniaki Saito, Kohei Watanabe, Yoshitaka Ushiku, and Tatsuya Harada.
\newblock Maximum classifier discrepancy for unsupervised domain adaptation.
\newblock In {\em Proceedings of the IEEE conference on computer vision and pattern recognition}, pages 3723--3732, 2018.

\bibitem{scholkopf2021toward}
Bernhard Sch{\"o}lkopf, Francesco Locatello, Stefan Bauer, Nan~Rosemary Ke, Nal Kalchbrenner, Anirudh Goyal, and Yoshua Bengio.
\newblock Toward causal representation learning.
\newblock {\em Proceedings of the IEEE}, 109(5):612--634, 2021.

\bibitem{schott2021visual}
Lukas Schott, Julius Von~K{\"u}gelgen, Frederik Tr{\"a}uble, Peter Gehler, Chris Russell, Matthias Bethge, Bernhard Sch{\"o}lkopf, Francesco Locatello, and Wieland Brendel.
\newblock Visual representation learning does not generalize strongly within the same domain.
\newblock {\em arXiv preprint arXiv:2107.08221}, 2021.

\bibitem{shah2020pitfalls}
Harshay Shah, Kaustav Tamuly, Aditi Raghunathan, Prateek Jain, and Praneeth Netrapalli.
\newblock The pitfalls of simplicity bias in neural networks.
\newblock {\em Advances in Neural Information Processing Systems}, 33:9573--9585, 2020.

\bibitem{sun2015return}
Baochen Sun, Jiashi Feng, and Kate Saenko.
\newblock Return of frustratingly easy domain adaptation. corr.
\newblock {\em arXiv preprint arXiv:1511.05547}, 2015.

\bibitem{szegedy2013intriguing}
Christian Szegedy, Wojciech Zaremba, Ilya Sutskever, Joan Bruna, Dumitru Erhan, Ian Goodfellow, and Rob Fergus.
\newblock Intriguing properties of neural networks.
\newblock {\em arXiv preprint arXiv:1312.6199}, 2013.

\bibitem{teney2022evading}
Damien Teney, Ehsan Abbasnejad, Simon Lucey, and Anton van~den Hengel.
\newblock Evading the simplicity bias: Training a diverse set of models discovers solutions with superior ood generalization.
\newblock In {\em Proceedings of the IEEE/CVF Conference on Computer Vision and Pattern Recognition}, pages 16761--16772, 2022.

\bibitem{teshima2020few}
Takeshi Teshima, Issei Sato, and Masashi Sugiyama.
\newblock Few-shot domain adaptation by causal mechanism transfer.
\newblock In {\em International Conference on Machine Learning}, pages 9458--9469. PMLR, 2020.

\bibitem{tzeng2017adversarial}
Eric Tzeng, Judy Hoffman, Kate Saenko, and Trevor Darrell.
\newblock Adversarial discriminative domain adaptation.
\newblock In {\em Proceedings of the IEEE conference on computer vision and pattern recognition}, pages 7167--7176, 2017.

\bibitem{van2008visualizing}
Laurens Van~der Maaten and Geoffrey Hinton.
\newblock Visualizing data using t-sne.
\newblock {\em Journal of machine learning research}, 9(11), 2008.

\bibitem{venkateswara2017deep}
Hemanth Venkateswara, Jose Eusebio, Shayok Chakraborty, and Sethuraman Panchanathan.
\newblock Deep hashing network for unsupervised domain adaptation.
\newblock In {\em Proceedings of the IEEE conference on computer vision and pattern recognition}, pages 5018--5027, 2017.

\bibitem{wang2021multimodal}
Luyu Wang, Pauline Luc, Adria Recasens, Jean-Baptiste Alayrac, and Aaron van~den Oord.
\newblock Multimodal self-supervised learning of general audio representations.
\newblock {\em arXiv preprint arXiv:2104.12807}, 2021.

\bibitem{wang2022out}
Xi~Wang and Laurence Aitchison.
\newblock Out of distribution robustness with pre-trained bayesian neural networks.
\newblock {\em arXiv preprint arXiv:2206.12361}, 2022.

\bibitem{wenzel2022assaying}
Florian Wenzel, Andrea Dittadi, Peter~Vincent Gehler, Carl-Johann Simon-Gabriel, Max Horn, Dominik Zietlow, David Kernert, Chris Russell, Thomas Brox, Bernt Schiele, et~al.
\newblock Assaying out-of-distribution generalization in transfer learning.
\newblock {\em arXiv preprint arXiv:2207.09239}, 2022.

\bibitem{wortsman2022model}
Mitchell Wortsman, Gabriel Ilharco, Samir~Ya Gadre, Rebecca Roelofs, Raphael Gontijo-Lopes, Ari~S Morcos, Hongseok Namkoong, Ali Farhadi, Yair Carmon, Simon Kornblith, et~al.
\newblock Model soups: averaging weights of multiple fine-tuned models improves accuracy without increasing inference time.
\newblock In {\em International Conference on Machine Learning}, pages 23965--23998. PMLR, 2022.

\bibitem{wu2022extrapolation}
Yongtao Wu, Zhenyu Zhu, Fanghui Liu, Grigorios~G Chrysos, and Volkan Cevher.
\newblock Extrapolation and spectral bias of neural nets with hadamard product: a polynomial net study.
\newblock {\em arXiv preprint arXiv:2209.07736}, 2022.

\bibitem{xu2020neural}
Keyulu Xu, Mozhi Zhang, Jingling Li, Simon~S Du, Ken-ichi Kawarabayashi, and Stefanie Jegelka.
\newblock How neural networks extrapolate: From feedforward to graph neural networks.
\newblock {\em arXiv preprint arXiv:2009.11848}, 2020.

\bibitem{xu2019d}
Xiang Xu, Xiong Zhou, Ragav Venkatesan, Gurumurthy Swaminathan, and Orchid Majumder.
\newblock d-sne: Domain adaptation using stochastic neighborhood embedding.
\newblock In {\em Proceedings of the IEEE/CVF Conference on Computer Vision and Pattern Recognition}, pages 2497--2506, 2019.

\bibitem{yalniz2019billion}
I~Zeki Yalniz, Herv{\'e} J{\'e}gou, Kan Chen, Manohar Paluri, and Dhruv Mahajan.
\newblock Billion-scale semi-supervised learning for image classification.
\newblock {\em arXiv preprint arXiv:1905.00546}, 2019.

\bibitem{yi2021improved}
Mingyang Yi, Lu~Hou, Jiacheng Sun, Lifeng Shang, Xin Jiang, Qun Liu, and Zhiming Ma.
\newblock Improved ood generalization via adversarial training and pretraing.
\newblock In {\em International Conference on Machine Learning}, pages 11987--11997. PMLR, 2021.

\bibitem{yu2022coca}
Jiahui Yu, Zirui Wang, Vijay Vasudevan, Legg Yeung, Mojtaba Seyedhosseini, and Yonghui Wu.
\newblock Coca: Contrastive captioners are image-text foundation models.
\newblock {\em arXiv preprint arXiv:2205.01917}, 2022.

\bibitem{zhang2021understanding}
Chiyuan Zhang, Samy Bengio, Moritz Hardt, Benjamin Recht, and Oriol Vinyals.
\newblock Understanding deep learning (still) requires rethinking generalization.
\newblock {\em Communications of the ACM}, 64(3):107--115, 2021.

\bibitem{zhang2017mixup}
Hongyi Zhang, Moustapha Cisse, Yann~N Dauphin, and David Lopez-Paz.
\newblock mixup: Beyond empirical risk minimization.
\newblock {\em arXiv preprint arXiv:1710.09412}, 2017.

\bibitem{zhang2022few}
Wenyu Zhang, Li~Shen, Wanyue Zhang, and Chuan-Sheng Foo.
\newblock Few-shot adaptation of pre-trained networks for domain shift.
\newblock {\em arXiv preprint arXiv:2205.15234}, 2022.

\end{thebibliography}

\end{document}